\documentclass[letterpaper, 10 pt, conference]{ieeeconf}  

\IEEEoverridecommandlockouts                              

\usepackage{graphics}
\usepackage{float} 
\usepackage{caption}
\usepackage{subfigure}
\usepackage{epsfig}
\usepackage{mathptmx}
\usepackage{times}
\usepackage{amsmath} 
\usepackage{amssymb}  
\usepackage{array}
\usepackage{url} 
\usepackage{color}
\usepackage{cite} 
\usepackage[ruled,linesnumbered]{algorithm2e}  
\usepackage{siunitx}
\usepackage{textcomp} 

\title{\LARGE \bf
	Frontier Detection and Reachability Analysis for Efficient 2D Graph-SLAM Based Active Exploration
}

\author{
	Zezhou Sun$^{\dag}$,
	Banghe Wu$^{\dag}$,
	Cheng-Zhong Xu$^{\ddag}$,
	Sanjay E. Sarma$^{\S}$,
	Jian Yang$^{\dag}$,
	and
	Hui Kong$^{\dag}$
	\thanks{$\dag$ School of Computer Science and Engineering, Nanjing University of Science and Technology, Nanjing, Jiangsu, China.}
	\thanks{$\ddag$ Department of Computer Science, University of Macau, Macau.}
	\thanks{$\S$ Department of Mechanical Engineering, MIT, Cambridge, MA.}
}

\begin{document}
	
	\maketitle
	\thispagestyle{empty}
	\pagestyle{empty}
	
	\begin{abstract} 
		
		We propose an integrated approach to active exploration by exploiting the Cartographer method as the base SLAM module for submap creation and performing efficient frontier detection in the geometrically co-aligned submaps induced by graph optimization. We also carry out analysis on the reachability of frontiers and their clusters to ensure that the detected frontier can be reached by robot. Our method is tested on a mobile robot in real indoor scene to demonstrate the effectiveness and efficiency of our approach. 
		
	\end{abstract}
	
	\section{INTRODUCTION}
	
	%
	
	
	Among all the existing active exploration works, Yamauchi et al's frontier-based method is a representative one \cite{yamauchi1997frontier}, where robot creates an occupancy grid map of the surrounding unknown environment through a SLAM algorithm and performs frontier detection in the grid map to find the rest of explorable areas.
	Frontier detection refers to finding the boundary between an unknown area and a known region that is not occupied by obstacles in a map. 
	
	The efficiency of frontier detection depends on the type of the selected SLAM algorithm. In general, either a filtering-based or a graph-based SLAM method can be exploited for mapping in an active exploration framework. 
	When using a filtering-based SLAM method, e.g., the Hector SLAM \cite{kohlbrecher2011flexible} and GMapping \cite{grisetti2007improved} based on Rao-Blackwellized particle filters (RBPF) \cite{doucet2000rao}, one can optimize the pose of the latest frame without changing those of the previous ones. Thus, the frontier detection step only needs to detect the frontiers induced by the latest frame, achieving a fast incremental frontier detection.
	In contrast, when using a graph-based SLAM method in active exploration, e.g., the Cartographer \cite{hess2016real}, each optimization step changes the poses of many (if not all) frames. Thus, one needs to re-detect all frontiers in many geometrically co-aligned frames, instead of only the ones in the latest frame. 
	
	Although frontier detection in a filtering-based SLAM active exploration is usually faster than that in a graph-based SLAM active exploration, a graph-based SLAM is generally more accurate than a filtering-based SLAM.
	Two recent articles \cite{filipenko2018comparison}  \cite{yagfarov2018map} evaluate  three representative methods (the Hector SLAM \cite{kohlbrecher2011flexible}, GMapping \cite{grisetti2007improved}, and Cartographer \cite{hess2016real}) in terms of mapping and trajectory accuracy, and conclude that the Cartographer method is better than the other two for 2D laser SLAM. In addition, according to \cite{hess2016real}, the scan-to-scan matching in the key-frame based graph-SLAM methods can quickly accumulate error, while the scan-to-map matching in Cartographer can reduce error accumulation. Therefore,  we choose Cartographer as our base SLAM module for more accurate mapping, although our method is applicable to a key-frame based (such as Pose SLAM \cite{ila2009information} and its variants) graph-SLAM active exploration.
	
	\begin{figure}[]
		\centering
		\includegraphics[width=1\linewidth]{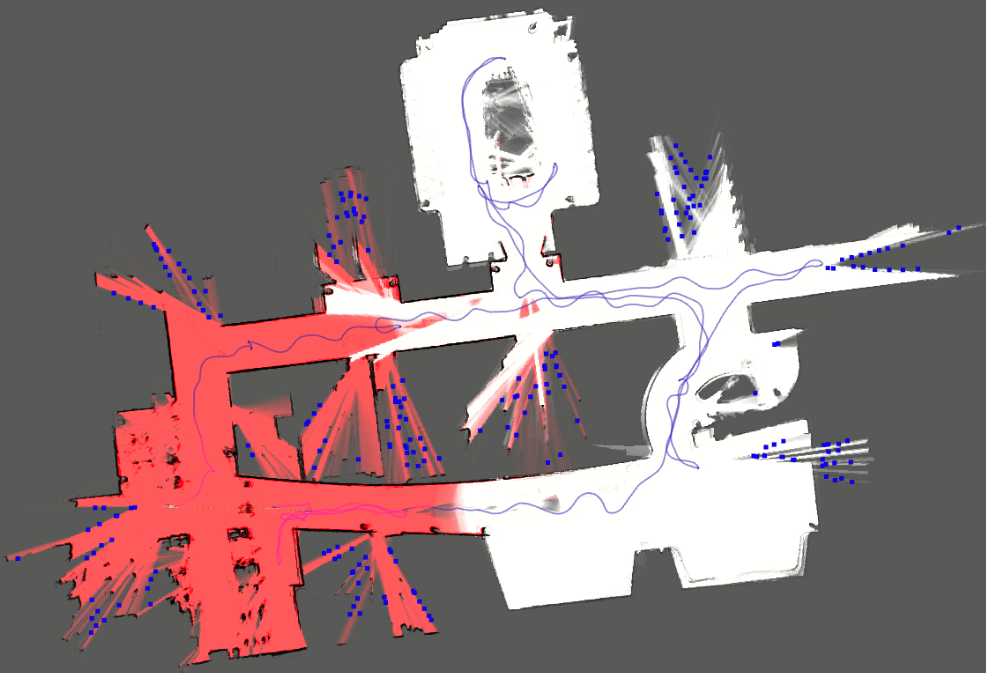}
		\caption{An office map created by a mobile robot using our method. The red part corresponds to the submaps where frontier detection runs and the blue points are the navigation points generated by our clustering method. The blue line is the exploration trajectory of the robot. }
		\label{globalmap}
	\end{figure}
	
	
	
	Recently, a couple of active exploration methods were proposed \cite{vallve2015active} \cite{valencia2018active} by exploiting graph-SLAM for mapping. 
	These methods iterates between mapping and frontier detection with evolving graph-optimization, and one needs to detect all frontiers in the updated global map. As map becomes larger, the time spent on frontier detection increases, making it hard to apply the existing frontier detection methods to large scenes. More recently, a dense frontier detection (DFD) method \cite{8760392} was proposed, where the local frontiers of each submap are detected and reserved. After the global map is updated by co-aligning each submap, the reserved frontiers are evaluated whether they are still frontiers in global map. However, the DFD method is still costly because all reserved frontiers should be evaluated. 
	
	In our work, instead of evaluating all submaps, we only select those submaps whose pose change is larger than a certain threhold after graph-optimization. By our method, the computation cost of frontier detection is only up to 60.7$\%$ of the DFD method in small scenes and 40.03$\%$ in large scenes. 
	In addition, some articles discuss whether the detected frontiers can be reached or not \cite{senarathne2015incremental} 
	\cite{li2016dynamic}. Frontiers can often be detected but cannot be reached due to narrow passage. Using these frontiers, robots could plan some invalid paths that can never be traversed, making active exploration very inefficient or even fail. Therefore, we analyze the reachability of local frontiers to ensure that the detected frontier must be reachable by robot. To make exploration efficient, clustering method is often used to reduce the number of frontiers. However, few articles notice that the commonly used clustering methods may undermine reachability of frontiers. Thus, we improve the clustering of frontiers to ensure that the obtained clusters of frontier are reachable. 
	
	The main contributions of this paper are as follows.
	1. We propose an integrated active exploration method for 2D graph SLAM based on Cartographer. 
	2. We improve the state-of-the-art frontier detection algorithm in the graph-SLAM induced map. Our method reduces the cost up to 60.7$\%$ of the state-of-the-art method, with an error rate up to no more than 1\textperthousand$ $ in our small dataset. In a public large dataset, our method reduces the cost up to 40.3$\%$ of the state-of-the-art method, with an error rate up to no more than 3\textperthousand$ $. 
	3. We analyze and improve reachability of frontiers, and obtain reachable navigation points by proposing a clustering method to obtain the clusters of dense frontiers as navigation points, thus avoiding the potentially invalid path planning due to the unreachable navigation points. 
	
	\section{Related works}
	
	Yamauchi et al. proposed their seminal active exploration work based on frontier detection and tracing \cite{yamauchi1997frontier}, where the grid map is divided into free, unknown, and occupied regions according to confidence, and the entire map is searched to find all frontiers. 
	
	Keidar proposed the Wavefront Frontier Detection (WFD) and Fast Frontier Detector (FFD) algorithms in \cite{keidar2012robot}. 
	The WFD runs two Breadth First Searches (BFS) on the map. 
	One searches outward from the robot's location, until it encounters the first unknown space. 
	The other starts from this unknown space, searching for continuous frontier. The WFD limits search space from whole map to free space. However, as the exploration area grows, one cannot expect the WFD to perform as fast as small scenes. 
	The FFD only processes new laser readings in real time. 
	It added frontier by processing adjacent scan points of the laser. 
	The FFD can perform real-time detection by processing raw laser data, rather than the free space of the map. 
	However, the FFD directly processes sensor data, so it cannot perform frontier detection on processed sensor information (such as expanding obstacles in the map or fusing multiple laser data). 
	
	The Expanding-Wavefront Frontier Detection (EWFD) \cite{quin2014expanding} combines the ideas of the WFD and FFD, which searches new unoccupied areas instead of searching all free spaces like the WFD. 
	Another kind of detection methods are called the Rapidly-exploring Random Trees (RRT) \cite{lavalle1998rapidly}.  
	The RRT detects sparse frontier points by randomly sampling and expanding in known space, such as \cite{umari2017autonomous}. 
	
	Some articles focus on generating safe, reachable frontier points. 
	\cite{wein2007visibility} extracts safe and reachable areas in the map through Voronoi diagrams. 
	\cite{senarathne2015incremental} proposed Safe and Reachable Frontier Detection (SRFD) method to incrementally manage the obstacle inflation. 
	It handles the change of the safe and reachable area caused by the addition and deletion of obstacles by processing the locally updated map data incrementally. 
	\cite{li2016dynamic} proposed a variant of SRFD called Safe and Reachable Frontier Detection Generator (SRFDG). 
	Different from the previous work, it first generate new coarse frontier points by processing the new laser data, then combine the previous frontier points and the global topology map to refine those new coarse frontier points, improving the efficiency of generating safe and reachable frontier points. 
	
	In addition, some clustering methods evaluate frontier point groups to improve exploration performance, such as the probability-density- \cite{umari2017autonomous}, histogram- \cite{mobarhani2011histogram} and semantic-information-based \cite{stachniss2006speeding} clustering methods. 
	
	\section{Prerequisites}\label{section:Prerequisites}
	
	\subsection{Definitions}
	
	\textbf{Occupancy labeling:} Our labeling of grid map is the same as Yamauchi \cite{yamauchi1997frontier}. 
	The $i$-th row and $j$-th column of the grid, $m_{i,j}$, is labeled according to its occupancy probability $p_{i,j}$. 
	The prior probability of each grid is set to 0.5. 
	
	\begin{equation}  
	Label(m_{i,j})=
	\left\{  
	\begin{aligned}
	&Free &p_{i,j} < 0.5 \\
	&Occupied  &p_{i,j} > 0.5 \\
	&Unknow  &p_{i,j} = 0.5 
	\end{aligned}  
	\right.  
	\end{equation}  
	
	\textbf{Frontier:} the set of all free grids adjacent to at least one unknown grid.
	
\begin{figure*}[ht]
		\subfigure[]{
			\includegraphics[width=5cm,height=3.5cm]{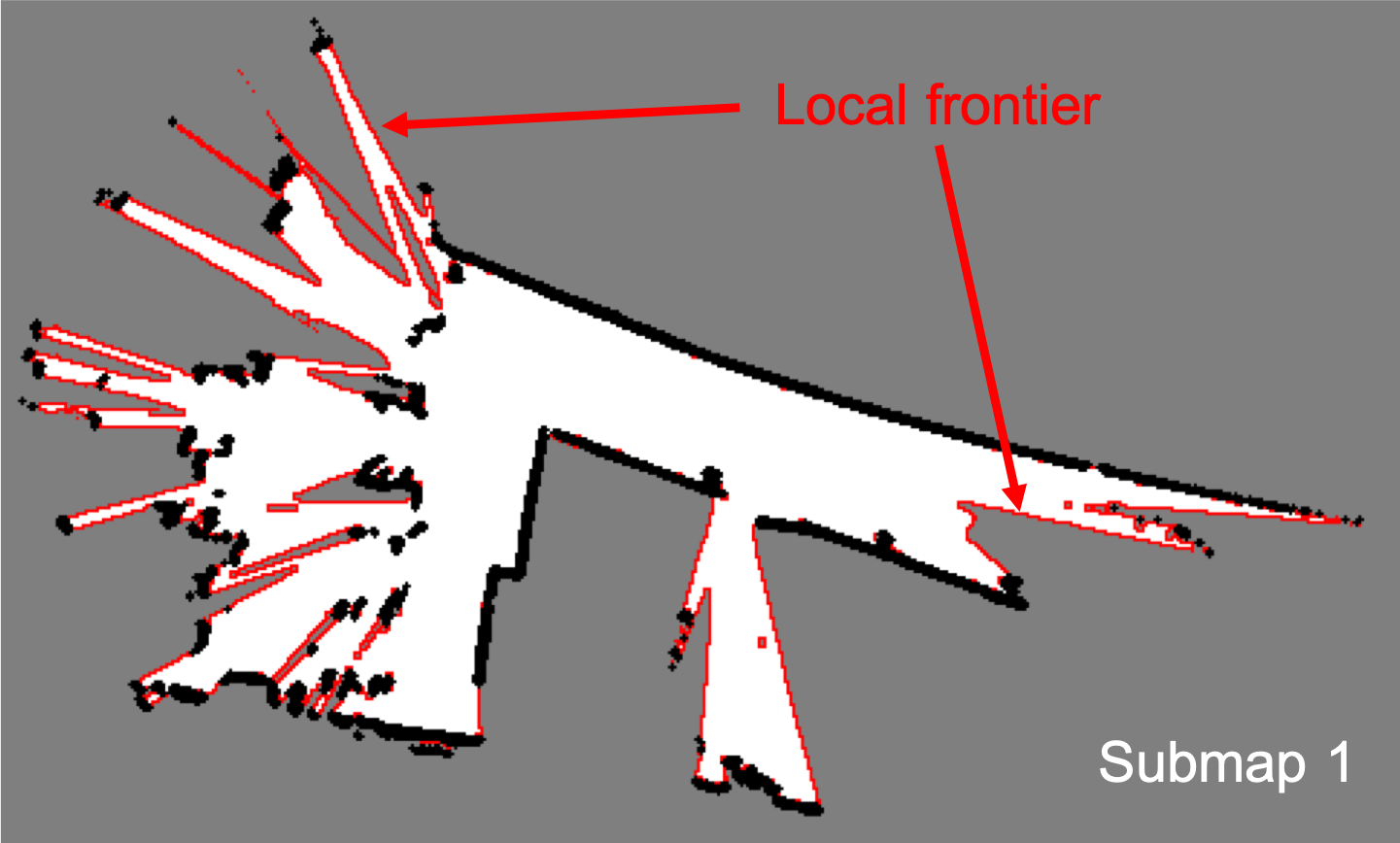}
		}
		\subfigure[]{
			\includegraphics[width=7cm,height=3.5cm]{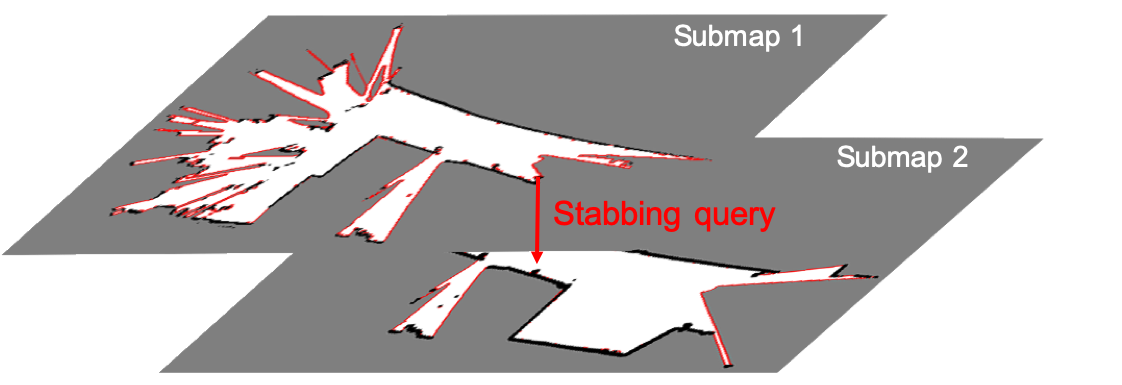}
		}
		\subfigure[]{
			\includegraphics[width=5cm,height=3.5cm]{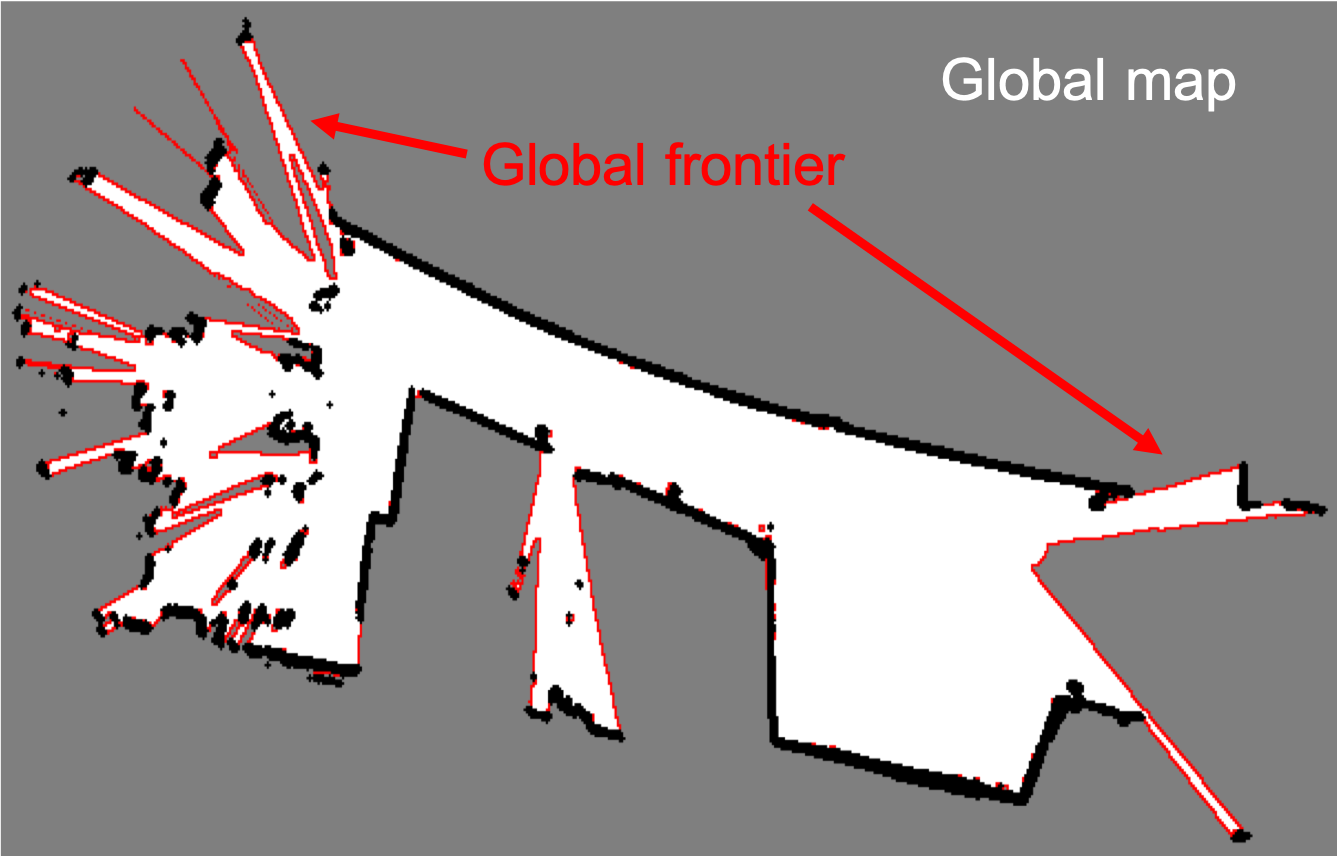}
		}
		\caption{(a) is a submap composed of several continuous laser scans. The red points are local frontiers detected by the WFD on the submap. 
			(b) shows how to perform stabbing query. According to the poses of the two submaps in the world coordinate system, we can calculate the corresponding position in submap $2$ of each local frontier point in submap $1$. 
			If the local frontier point in submap $1$ is still a local frontier point or an unknown point in submap $2$, we consider this frontier point to be a global frontier point (e.g., the global map is only composed of  two submaps). 
			Otherwise it indicates that this frontier point is occluded in submap $2$. 
			We do the same query operation for the local frontier points in submap $2$. 
			After we query all the local frontier points of the two submaps, we find out the global frontier of the global map, as shown in (c). }
		\label{defination}
	\end{figure*}

	\textbf{Submap:} a small occupancy grid map composed of several consecutive laser scans (see Fig.\ref{defination}(a)). 
	The poses of laser scans and frontier points contained in each submap are stored in a local coordinate system relative to the submap.
	
	\textbf{Global map:} a fused occupancy grid map constructed by joining all submaps according to the relative pose obtained from the graph optimization (see Fig.\ref{defination}(c)). 
	Since the relative pose between submaps changes after each optimization, the global map needs to be regenerated after each optimization. 
	
	\textbf{Local frontier:} the set of frontier points belonging to each submap (see Fig.\ref{defination}(a)). 
	
	\textbf{Global frontier:} the set of all frontier points of the global map (see Fig.\ref{defination}(c)). 
	
	\textbf{Stabbing query:} an operation used to evaluate whether a local frontier is a global one (Fig.\ref{defination}(b)). 
	A local frontier points are considered as a global one if and only if it belongs to frontier or unknown state in all geometrically co-aligned submaps. 
	
	\subsection{Cartographer}
	
	Cartographer is an open-sourced graph SLAM algorithm developed by Google \cite{hess2016real}. 
	It builds submaps and optimizes the pose of all scans and submaps following Sparse Pose Adjustment (SPA) \cite{konolige2010efficient}. It is supposed that the cumulative error in each submap is sufficiently small. 
	Once the submap is constructed, graph optimization only optimizes the relative poses between the submaps and does not change the relative poses of the laser scans inside submaps. In our work, we set the map resolution to $5$cm and each submap holds 70 laser scans. 
	
	\subsection{Dense Frontier Detection}
	
	The DFD \cite{8760392} implements a real-time dense frontier detection algorithm embedded in graph SLAM based on the fact that the relative poses of laser scans in each submap remain unchanged. 
	It detects dense frontiers in each submap once it is generated. 
	These local frontiers are not affected by graph optimization. 
	Then after each graph optimization, it performs stabbing query on the local frontiers of all submaps. 
	To speed up the query, the stabbing query only checks the submaps whose bounding boxes intersect with each other. 
	
	\section{Our Methods}
	
	\subsection{Reachability of Frontiers}
	
\begin{figure}[h]
		\subfigure[]{
			\includegraphics[width=4cm,height=3cm]{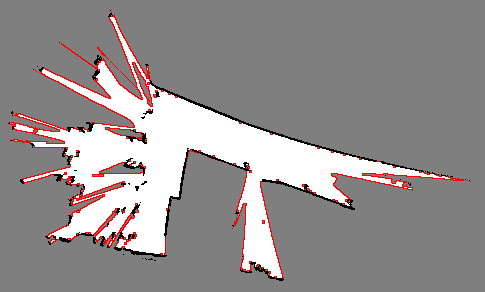}
		}
		\subfigure[]{
			\includegraphics[width=4cm,height=3cm]{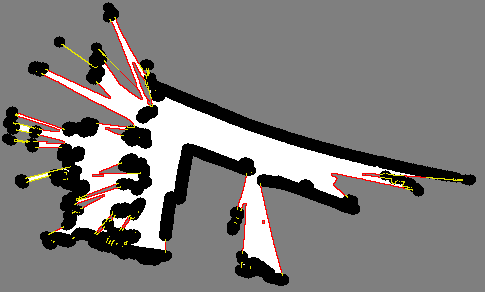}
		}
		\caption{(a) and (b) are submaps created by Cartographer before and after inflation, respectively. 
			The red points are the detected frontiers by the WFD method, and black color represents occupied regions. 
			It can be seen that there are many unreachable frontier points (yellow points in (b)) in (a) when the submap is not inflated. 
			Detecting and trying to reach these frontier points are meaningless and wasteful. 
			After we dilate the obstacle (see (b)), the unreachable frontiers cannot be detected by the WFD because of connectivity.}
		\label{inflate}
	\end{figure}

	Previous works detect safe and reachable frontiers on a global map or global topology map.  However, we only need to detect frontiers of the submaps. 
	By Cartographer, cumulative error in each submap is negligible, and reachable frontiers detected in each submap are mostly still reachable after fusing submaps.
	Therefore, different from the DFD method, we first dilate the submap and then use the WFD to detect its frontier. The purpose is not only to detect local frontier, but to detect local reachable frontier, as shown in Fig.\ref{inflate}. 
	The inflation operation on the submap ensures that the robot can reach the detected frontier. 
	The expansion radius should exceed that of the robot platform to ensure that the detected frontiers are reachable. For safety, we set an additional safety distance to avoid collision. In our work, because our robot platform has a radius of $20cm$ and we always expect the robot to keep a distance of $20cm$ away from obstacles, we set the inflation radius to $8$ grids ($40$cm). 
	
	\subsection{Frontier Detection}
	
	
	The DFD method re-transforms and re-tests local frontiers of all submaps. In contrast, we perform frontier detection on those submaps with significant deviations caused by optimization. 
	As shown in Algorithm \ref{algorithm1}, $PP$ and $CP$ represent the poses of the submaps before and after optimization, respectively. 
	$PF$ and $CF$ represent the global frontiers before and after optimization, respectively. 
	We use the subscript $i$ to represent the i-th submap and save the bounding-box of each submap in $global\_submap\_bounding\_box$.  
	In line 3, for any submap $N$, $global\_submap\_bounding\_boxes.Intersect$($N$) finds all submaps that intersect with $N$ according to the saved bounding-box. 
	We perform the BFS starting from the latest inserted submap, $N$, and search all submaps which intersect with $N$. 
	If the pose changes of these selected submaps exceed a threshold, we push these submaps into the BFS queue and stabbing-query queue (line 1-9). 
	The poses of submaps $CP_{i}$ and $PP_{i}$ indicate their three parameters $x, y$ and $\theta$ (line 4), respectively. 
	DeviationExceedsThreshold($CP_{i},PP_{i},\epsilon$) is $True$ when
	\begin{equation}
	\left\{
	\begin{aligned}
	&abs(CP_{i}.x-PP_{i}.x)>\epsilon \quad or \\
	&abs(CP_{i}.y-PP_{i}.y)>\epsilon \quad or \\
	&abs(CP_{i}.\theta-PP_{i}.\theta)>\frac{\epsilon}{2\pi R}
	\end{aligned}
	\right.
	\end{equation}
	where $R$, set to 8m, refers to our lidar's detection range. 
	Then we execute stabbing query on the submaps, $S_i$, whose pose change exceeds the threshold and the ones that intersect with $S_i$ (line 10-13). 
	For the submaps that do not exceed the threshold, we  use the previously detected frontier instead (line 14-16). 
	The frontier's position we store are relative to the submap. 
	Therefore, the coordinates of the previously detected frontier can be calculated directly from optimization-corrected submaps without introducing errors.  
	Empirically, the threshold $\epsilon$ is set to $5$cm (one pixel). 
	
	\begin{algorithm}[ht]
		\caption{The BFS method}
		\label{algorithm1}
		\KwIn{ \\
			\quad submaps\_current\_pose: $CP$ \\
			\quad submaps\_previous\_pose: $PP$ \\
			\quad submaps\_previous\_frontier: $PF$ \\
			\quad global\_submap\_bounding\_boxes \\
			\quad BFS\_queue $\leftarrow$ latest submap \\
			\quad stabbing\_query\_queue $\leftarrow$ latest submap }
		\KwOut{\\
			\quad submaps\_current\_frontier $CF$ }
		\While{BFS\_queue is not empty}{ N $\leftarrow$ POP(BFS\_queue) \\
			\ForEach{$S_{i}\in$ global\_submap\_bounding\_boxes.Intersect($N$)}{
				\If{DeviationExceedsThreshold($CP_{i}$,$PP_{i}$,$\epsilon$) is True}{
					stabbing\_query\_queue $\leftarrow$ $S_{i}$ \\
					BFS\_queue $\leftarrow$ $S_{i}$
				}
			}
		}
		\ForEach{ $S_{i}\in$ stabbing\_query\_queue}{
			stabbing\_query\_queue $\leftarrow$ global\_submap\_bounding\_boxes.Intersect($S_{i}$)\\
			$CF\leftarrow$StabbingQuery($S_{i}$)
		}
		\ForEach{$S_{i}\notin$ stabbing\_query\_queue}{
			$CF\leftarrow PF_{i}$
		}
	\end{algorithm}
	
	We notice that the BFS method only considers pose change caused by single-round optimization and ignores accumulative pose change induced by multiple-round optimization. Thus,
	to deal with this issue, we directly record the accumulative pose changes of all submaps after optimization each round and execute frontier detection algorithm on the submaps $S_i$ whose accumulative change exceed the threshold and the ones that intersect with $S_i$ (line 1-11), as shown in Algorithm \ref{algorithm2}. 
	$CD$ is used to store the accumulative pose change of all submaps. 
	In line 2, $\oplus$ and $\ominus$ represent the plus and minus operation of the three parameters $x, y$ and $\theta$, respectively. 
	
	\begin{equation}
	CD_{i}\oplus CP_{i}\ominus PP_{i}=\left\{
	\begin{aligned}
	&CD_{i}.x+CP_{i}.x-PP_{i}.x \\
	&CD_{i}.y+CP_{i}.y-PP_{i}.y \\
	&CD_{i}.\theta+CP_{i}.\theta-PP_{i}.\theta \\
	\end{aligned}
	\right.
	\end{equation}
	After stabbing query, we reset the accumulative pose change of submaps  to $0$ (line 10). 
	
	\begin{algorithm}[ht]
		\caption{The Direct method}
		\label{algorithm2}
		\KwIn{ \\
			\quad submaps\_current\_pose $CP$ \\
			\quad submaps\_previous\_pose $PP$ \\
			\quad submaps\_previous\_frontier $PF$ \\
			\quad submaps\_cumulative\_deviation $CD$ \\
			\quad global\_submap\_bounding\_boxes \\
			\quad stabbing\_query\_queue $\leftarrow$ empty }
		\KwOut{ \\
			\quad submaps\_current\_frontier $CF$ }
		\ForEach{submap $S_{i}$}{
			$CD_{i} = CD_{i}\oplus CP_{i}\ominus PP_{i}$ \\
			\If{DeviationExceedsThreshold($CD_{i}$,$\epsilon$) is True}{
				stabbing\_query\_queue $\leftarrow$ $S_{i}$
			}
		}
		\ForEach{ $S_{i}\in$ stabbing\_query\_queue}{
			stabbing\_query\_queue $\leftarrow$ global\_submap\_bounding\_boxes.Intersect($S_{i}$)\\
			$CF \leftarrow$ StabbingQuery($S_{i}$) \\
			$CD_{i} = 0$
		}
		\ForEach{$S_{i}\notin$ stabbing\_query\_queue}{
			$CF\leftarrow PF_{i}$
		}
	\end{algorithm}
	
\begin{figure}[ht]
		\subfigure[]{
			\includegraphics[width=2.5cm,height=2.5cm]{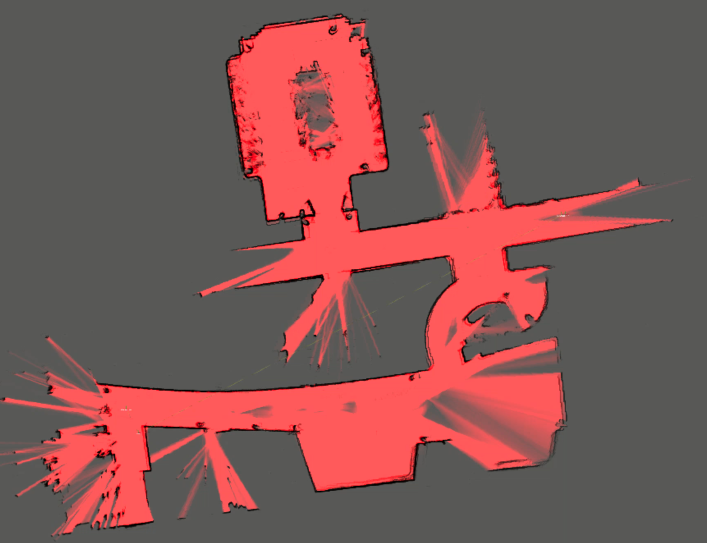}
		}
		\subfigure[]{
			\includegraphics[width=2.5cm,height=2.5cm]{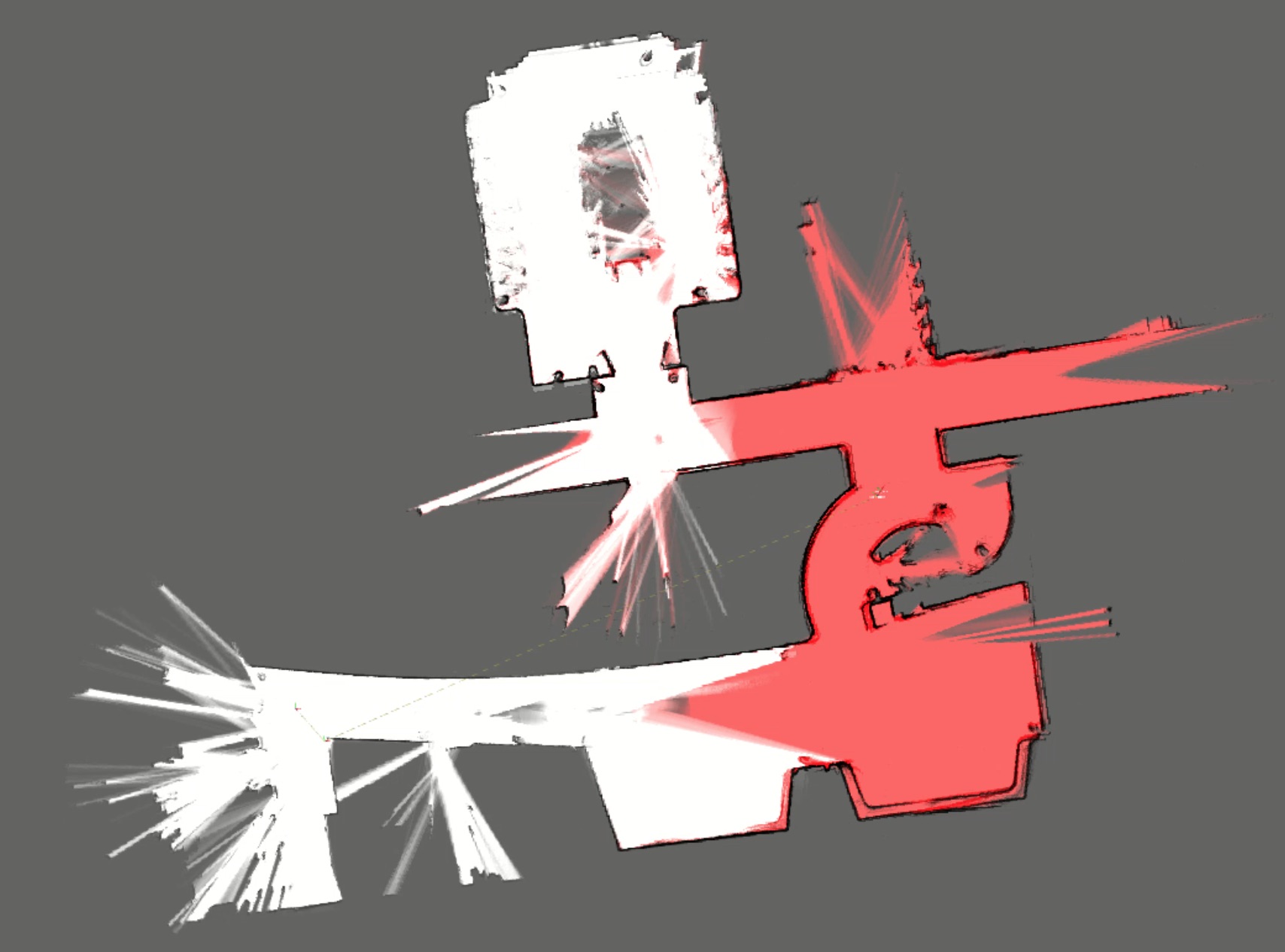}
		}
		\subfigure[]{
			\includegraphics[width=2.5cm,height=2.5cm]{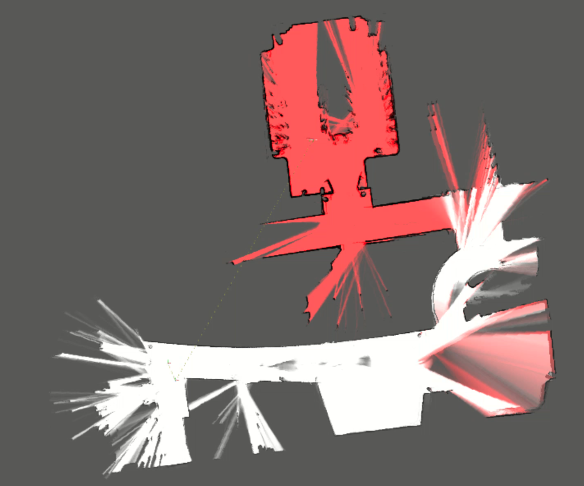}
		}
		\caption{We demonstrate the span of frontier detection (shown in red) in the DFD, the BFS and the Direct methods, respectively. 
			The white parts indicate the areas where frontiers keep unchanged from last round. 
			(a) the detection span of the DFD method. 
			After each round of optimization, all submaps are involved for frontier detection. 
			(b) shows the detection span of the BFS method. 
			The frontier detection starts from the latest submap until the pose change of all intersecting submaps is less than a threshold. 
			It ignores the cumulative pose change. 
			(c) shows the detection span of the Direct method. 
			It runs frontier detection in the submaps $S_i$ whose accumulative shift are larger than one pixel and the submaps that intersect with $S_i$. 
			Compared with the BFS method, it can detect submaps whose accumulative shifts exceed the threshold induced by multiple-round optimization. }
		\label{adaptive}
	\end{figure}

	When the optimization process induces negligible pose change of submaps, frontier detection is usually triggered only in a small range. On the contrary, 
	if the optimization process arouses significantly change in the pose of submaps (such as during a loop closure), frontier detection usually spans a large range of the map. Compared with the frontier detection of all submaps proposed by the DFD method, our method adaptively selects the span according to pose deviation of submaps, which makes frontier detection more efficient and flexible than the DFD method, as shown in Fig.\ref{adaptive}.  
	
	It should be emphasized that our method of adaptively selecting the span of frontier detection is not only applicable to Cartographer, but also to all key-frames based SLAM, such as active Pose SLAM \cite{valencia2018active}. To do that, we only need to replace the local frontier detection for the submaps with that for the key-frame scans. 
	
	\subsection{Clustering Dense Frontiers into Navigation Points}
	
	The detected frontiers are mostly continuous and dense, and they are usually 
	redundant for either path-planning or navigation purposes. Therefore, 
	a clustering operation on the detected dense frontiers is applied to sparsify the frontiers. This can reduce the computational burden of sorting the priority of frontiers. 
	Robot can choose some representative clustered points and set them as targets for exploration. Thus, we call these representative clustered points navigation points. 
	
	Generally, it is not robust to use random sampling or equi-distance selection of frontiers to generate navigation points, because this can miss small frontier segments. 
	Most active exploration methods use basic clustering algorithms. 
	For example, \cite{umari2017autonomous} uses the Mean-shift algorithm \cite{comaniciu2002mean} and \cite{puig2011new} uses the K-Means method. 
	However, previous works barely noticed that the exiting clustering methods may result in un-reachable frontiers. 
	To deal with this issue, we make two improvements to clustering. 
	
\begin{figure}[ht]
		\subfigure[]{
			\includegraphics[width=2.5cm,height=2.5cm]{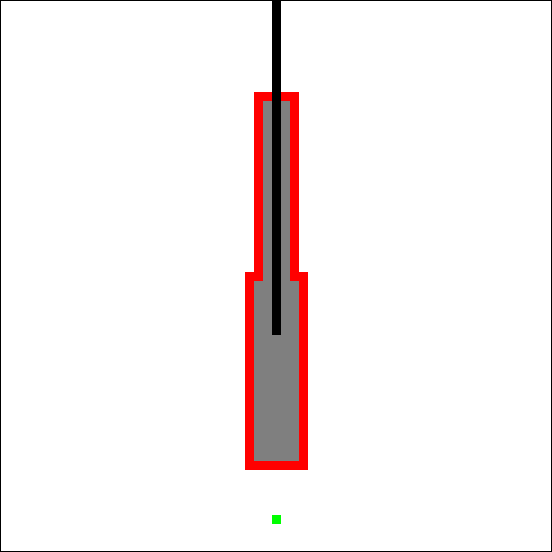}
		}
		\subfigure[]{
			\includegraphics[width=2.5cm,height=2.5cm]{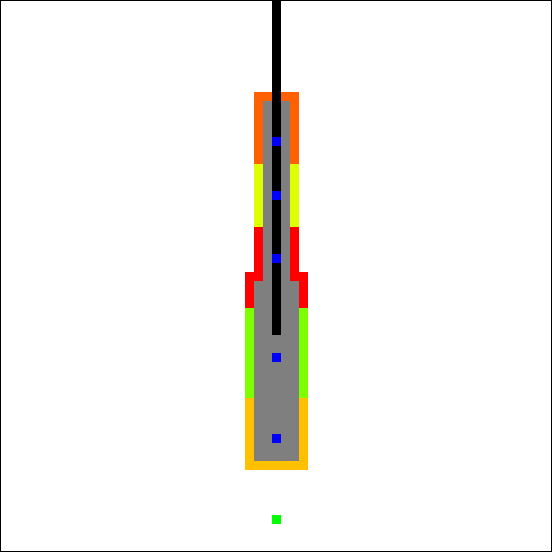}
		}
		\subfigure[]{
			\includegraphics[width=2.5cm,height=2.5cm]{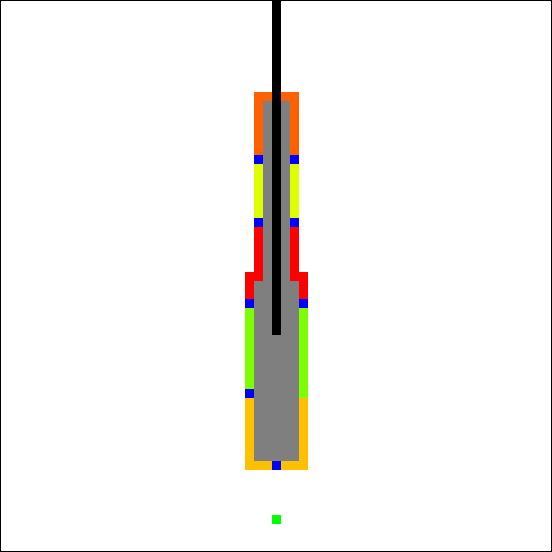}
		}
		\caption{(a) shows the dense frontier points on the simulated map. The white part represents free space, the black part representing obstacle, the gray part representing the unknown area, the green point being the robot, and the red part being the dense frontiers. 
			(b) shows the clustering results by applying the Mean-shift on the simulated map. The same color represents the same cluster. The blue points represent clustering results. 
			(c) shows the clustering results of our method on this map. It can be seen that the disconnected frontiers are clustered into different clusters, and the navigation points are ensured to be reachable. }
		\label{simulated}
	\end{figure}

\begin{figure}[h]
		\subfigure[]{
			\includegraphics[width=2.5cm,height=2.5cm]{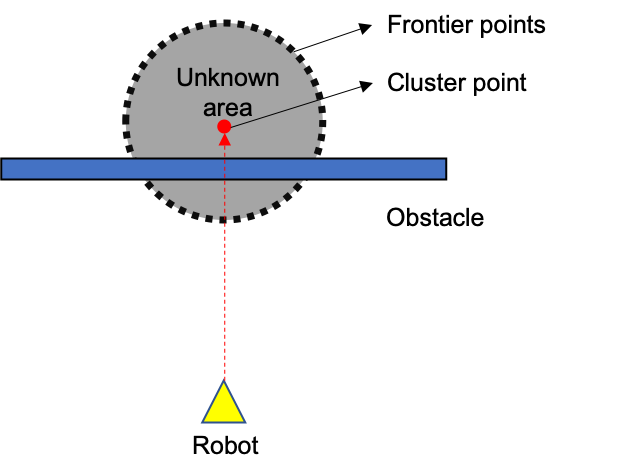}
		}
		\subfigure[]{
			\includegraphics[width=2.5cm,height=2.5cm]{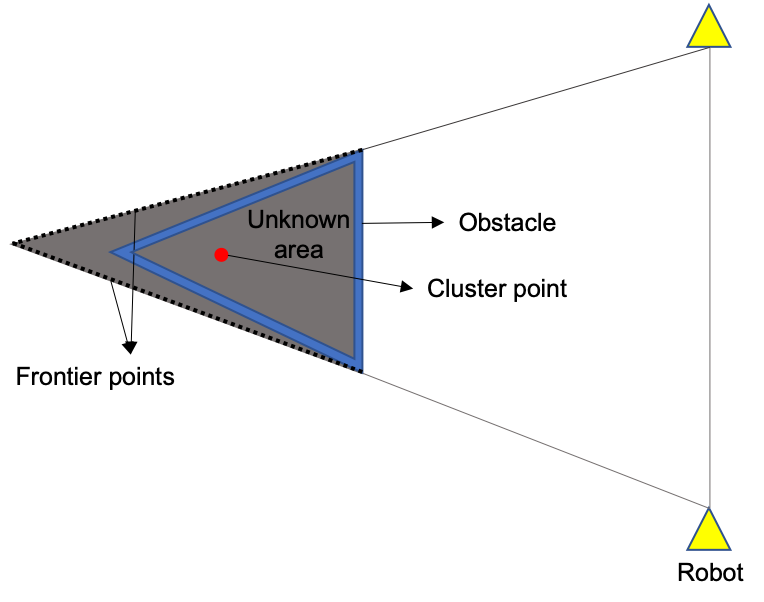}
		}
		\subfigure[]{
			\includegraphics[width=2.5cm,height=2.5cm]{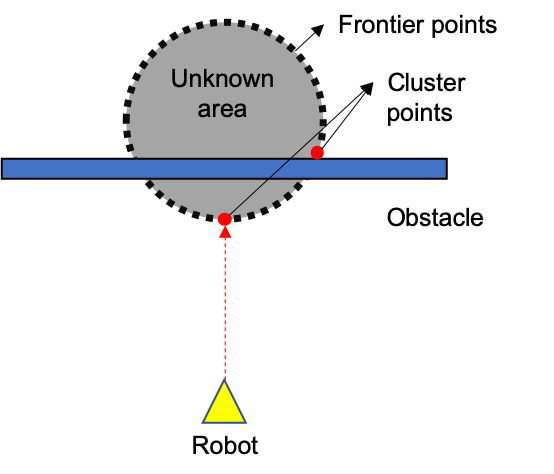}
		}
		\caption{(a) the existing clustering methods can easily group the disconnected frontiers into the same cluster, making the clustering center unreachable. 
			(b) If the clustering centers are set as the navigation points, they cannot be reached or observed either. 
			(c) We set the frontier point closest to the robot as the navigation point of each cluster of connected frontiers.}
		\label{clustererr}
	\end{figure}

	First, we notice that connectivity of frontier points is not taken into account by the existing frontier clustering methods. 
	Disconnected frontiers make clusters difficult to reach, as shown in Fig.\ref{clustererr}(a). 
	Fig.\ref{simulated}(a) shows the dense frontier of the simulated map. 
	We try the Mean-shift method to cluster the dense frontier of the simulated map  (see Fig.\ref{simulated}(b)). 
	The frontier points of the same color represent that they are in the same cluster. 
	It can be seen that some clusters contain disconnected frontiers. Mean-shift incorrectly clusters them into one and sets the location of clusters inside obstacle. 
	
	Due to the large number of frontiers, it is costly to check the connectivity of these points.
	Since our goal is to robustly disperse the dense frontier points instead of only clustering them. 
	Thus, we check the connectivity of frontier points within each cluster generated by Mean-shift. 
	We set both the unknown and known areas connected and only the occupied region can cause disconnection (see Fig.\ref{simulated}(c)). 
	If all the frontier points are connected, they can be clustered into one point. 
	If there are multiple connected areas, each connected area is clustered into a point. 
	As shown in Fig.\ref{simulated}(c), the dense frontiers are correctly clustered and sparsified in terms of connectivity. The clusters obtained in this way not only conform to human intuition, but also facilitate robots to generate the exploration path. 
	
	Second, instead of setting the navigation point as the center of each cluster (see Fig.\ref{simulated}(b) and Fig.\ref{clustererr}(b)), we set the frontier point closest to the robot in the connected area as the navigation point of each cluster of connected frontiers, as shown in Fig.\ref{simulated}(c) and Fig.\ref{clustererr}(c). 
	In this way, the navigation points are ensured to be safe and reachable.

	After getting the navigation points, we sort them according to the distance from the robot and the proportion of the unknown area caround the point, which is the same as \cite{umari2017autonomous}. 
	We assign the navigation points with the highest priority to the robot. The robot drives to the target through the move-base module of ROS. 
	Once the robot reaches the target or if the path planning from the robot to the target fails, the navigation point is updated in the sorted order. 
	
	\section{Experimental Results}
	
	\subsection{Mobile Robot Platform}
	
\begin{figure}[h]
		\centering
		\includegraphics[width=1\linewidth]{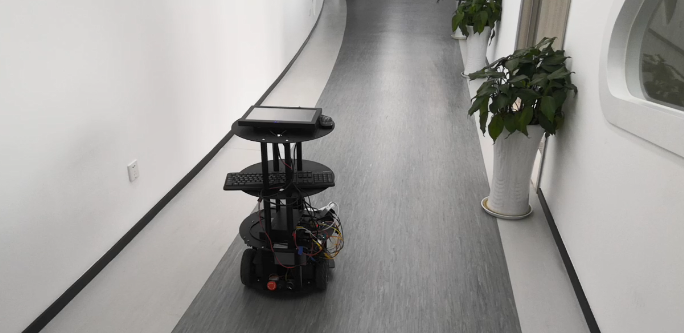}
		\caption{The E2-robot platform used for experiments. }
		\label{E2robot}
	\end{figure}

	Our experiment platform is an E2-robot, as shown in Fig.\ref{E2robot}. 
	The robot is driven differentially by rear wheels and front wheels are auxiliary. 
	It is equipped with a Hokuyo laser scanner ($8$m range and \ang{180} field of view), an IMU sensor, and an on-board computer with an Intel i7 processor and 32GB RAM running ROS. 
	
	\subsection{Frontier Detection Results} 
	
\begin{figure*}[h]
		\centering
		\subfigure[Explore near cluttered office items]{
			\includegraphics[width=5.5cm,height=3.6cm]{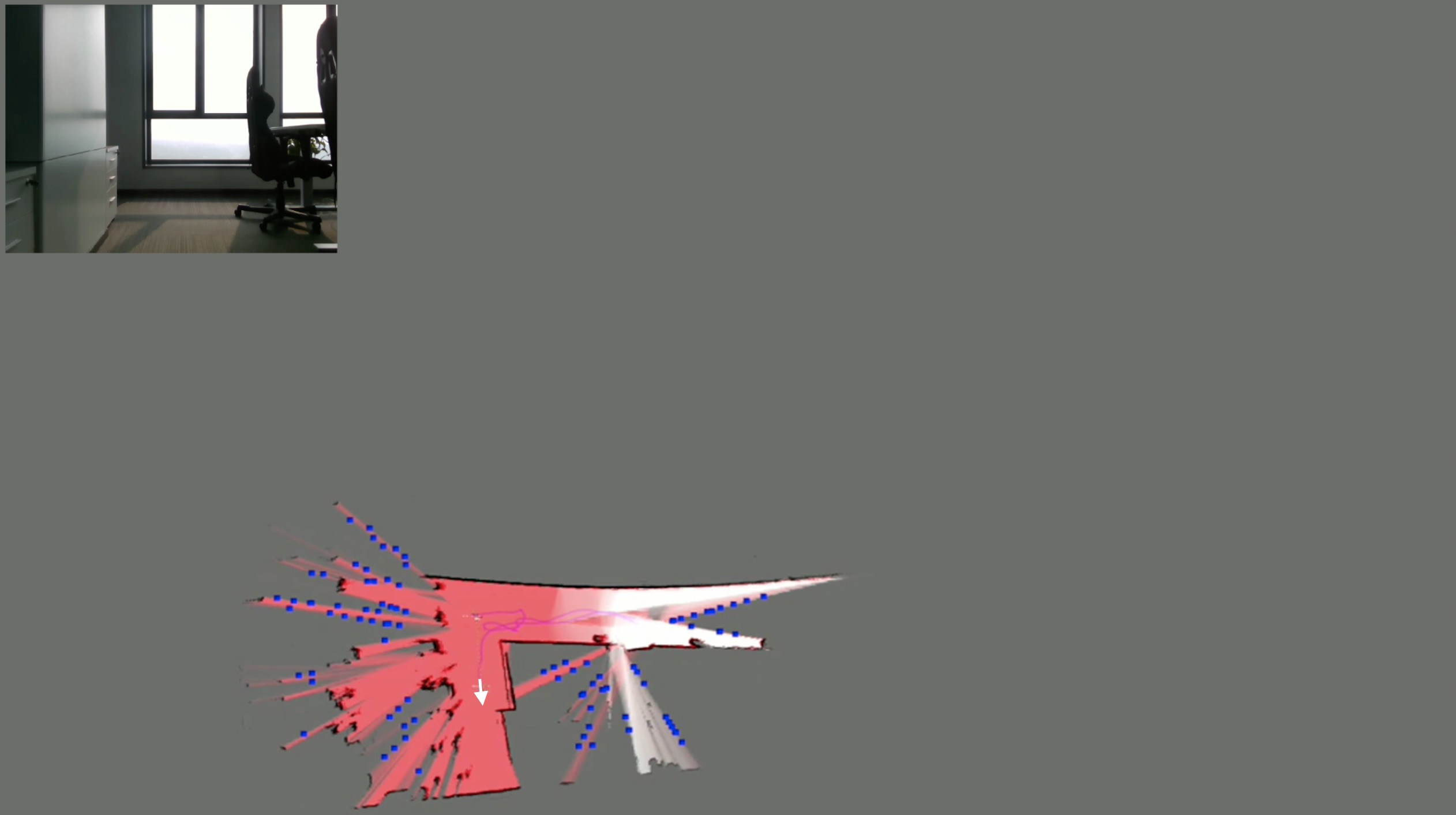}
		}
		\subfigure[Explore the corridor]{
			\includegraphics[width=5.5cm,height=3.6cm]{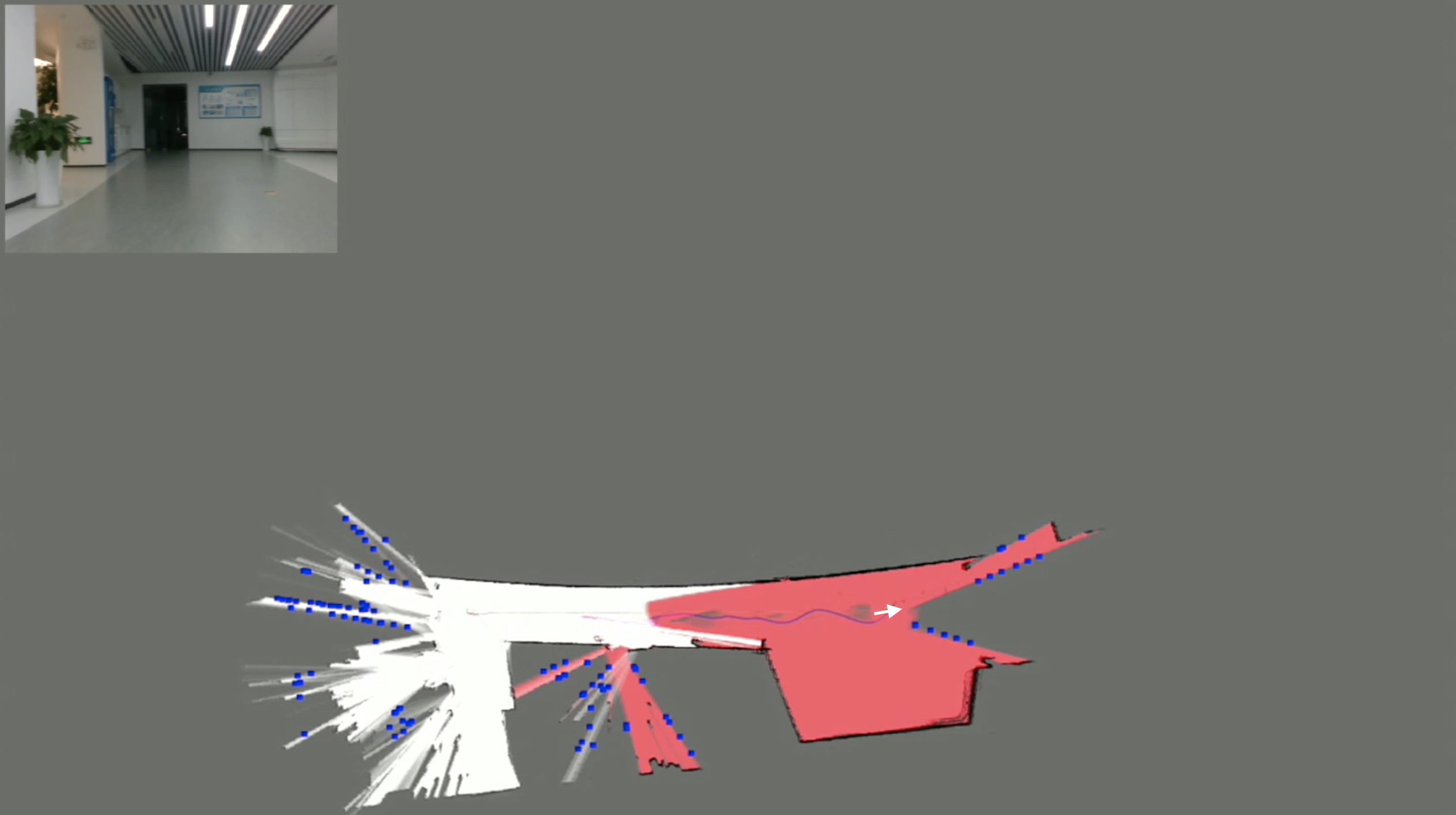}
		}
		\subfigure[Explore the meeting room]{
			\includegraphics[width=5.5cm,height=3.6cm]{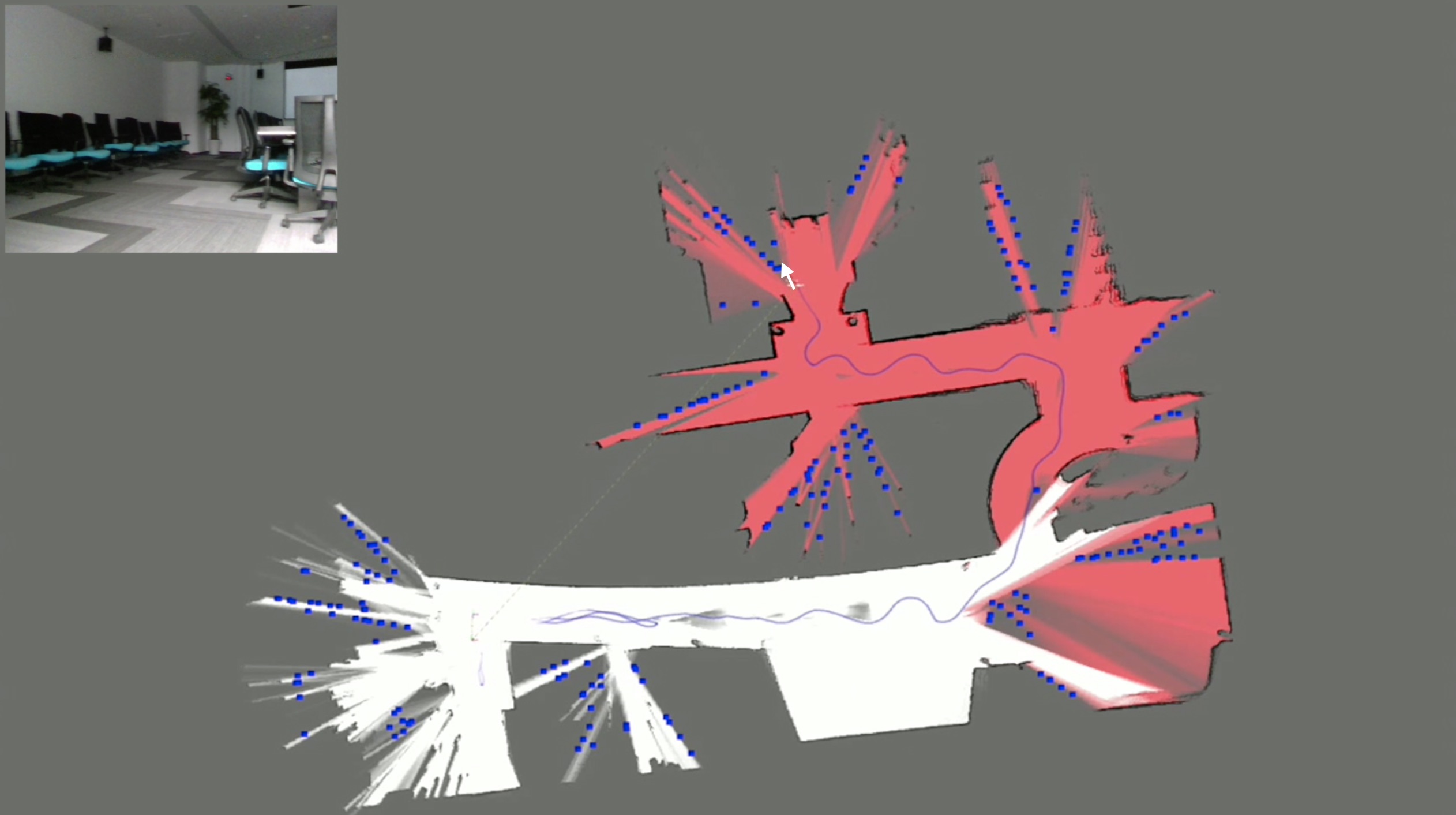}
		} 
		\subfigure[Small loop closure detected]{
			\includegraphics[width=5.5cm,height=3.6cm]{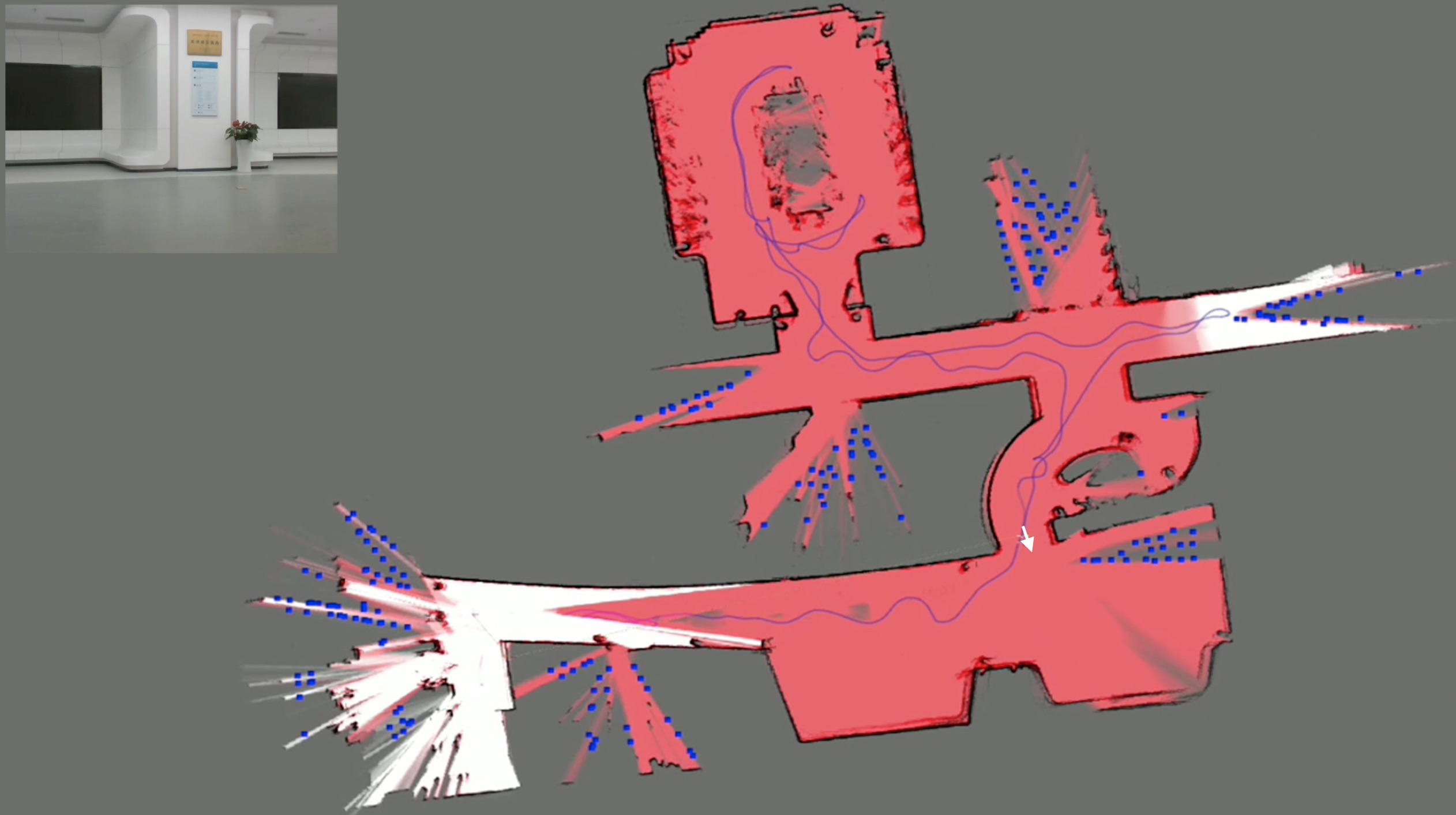}
		}
		\subfigure[Large loop closure detected]{
			\includegraphics[width=5.5cm,height=3.6cm]{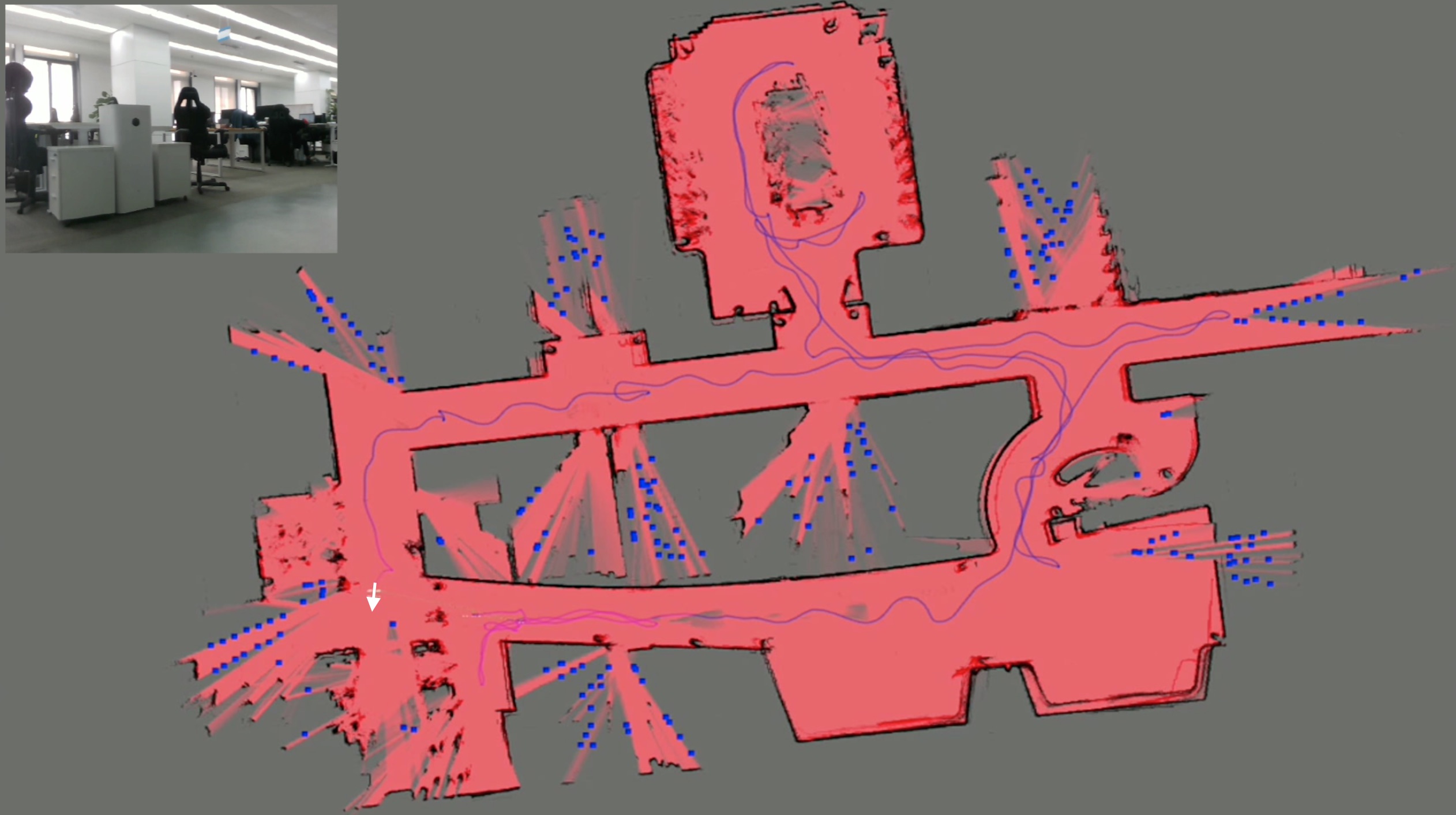}
		}
		\subfigure[Explore new areas after the loop closure]{
			\includegraphics[width=5.5cm,height=3.6cm]{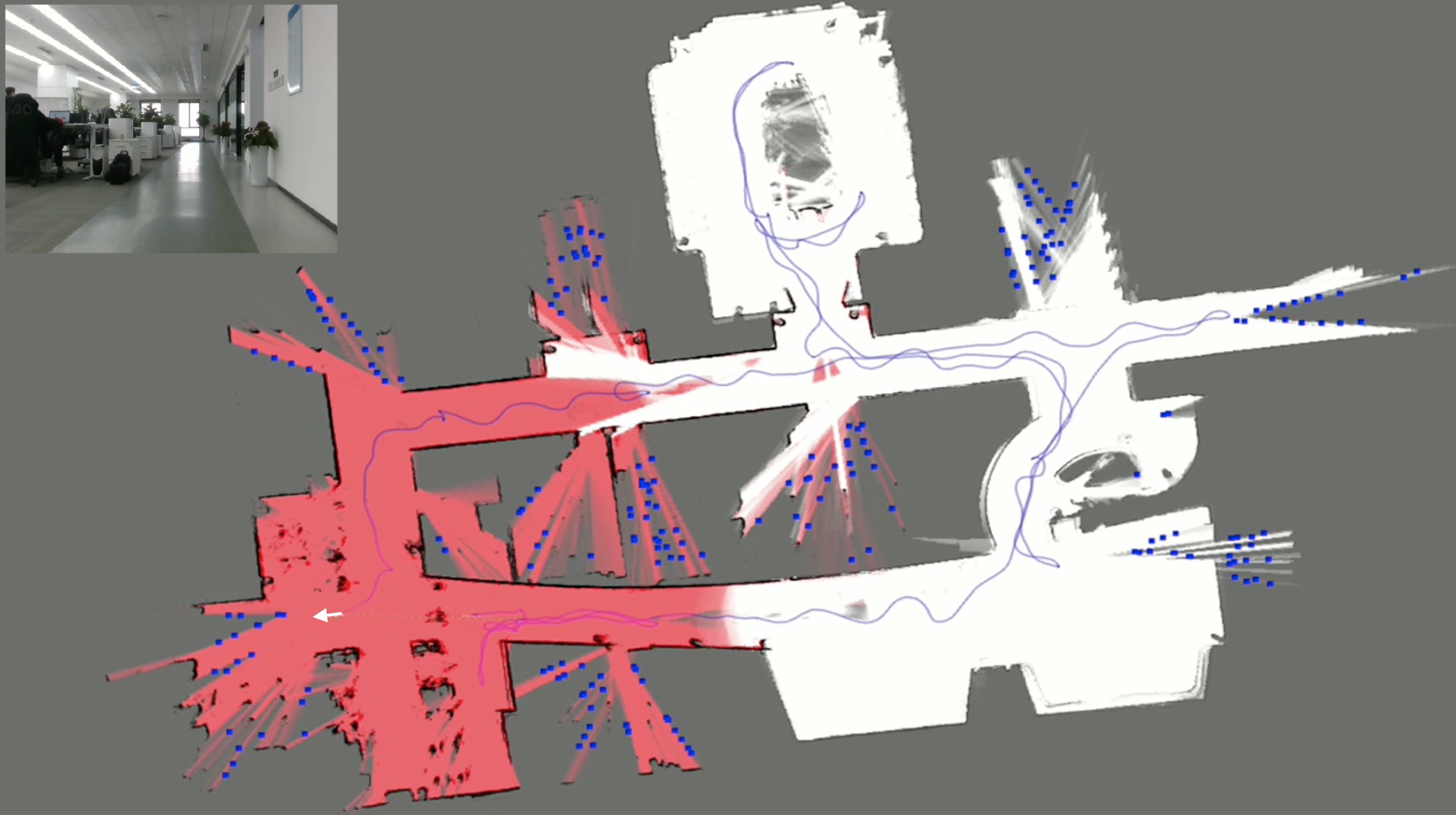}
		}
		\caption{(a-f) show the active exploration map of the robot at different moments in the office scene. 
			The white arrow in each figure represents the current position and orientation of the robot. 
			The image in the upper left corner shows how the office scene looks from the perspective of the current robot position. 
			The blue points represent the generated navigation points. The red part shows the span of frontier detection required by our method at the moment, and the blue line shows the robot's exploration trajectory. }
		\label{demo}
	\end{figure*}

	As shown in Fig.\ref{demo}, we show some moments of our robot's active exploration process. 
	Our robot is not disturbed by cluttered obstacles (tables, chairs or other office items), and can explore real office scenes robustly and autonomously. 
	Based on our method, the span of frontier detection (red part) is always, except when a large closed loop is found, smaller than that of the DFD method (all exploration areas). 
	
\begin{figure*}[h]
		\subfigure[Office scene]{
			\includegraphics[width=3.9cm,height=3cm]{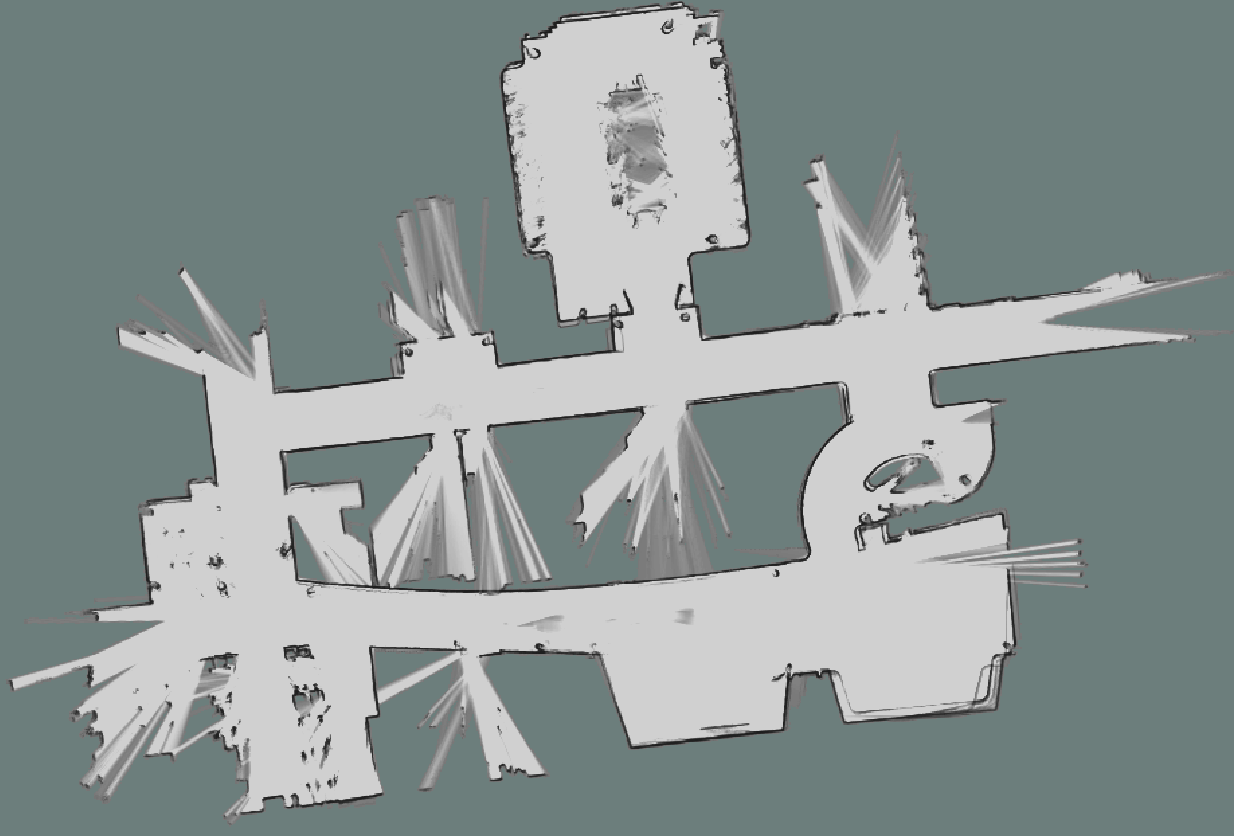}
		}\quad
		\subfigure[Total local frontier points of all submaps in the office scene]{
			\includegraphics[width=3.9cm,height=3cm]{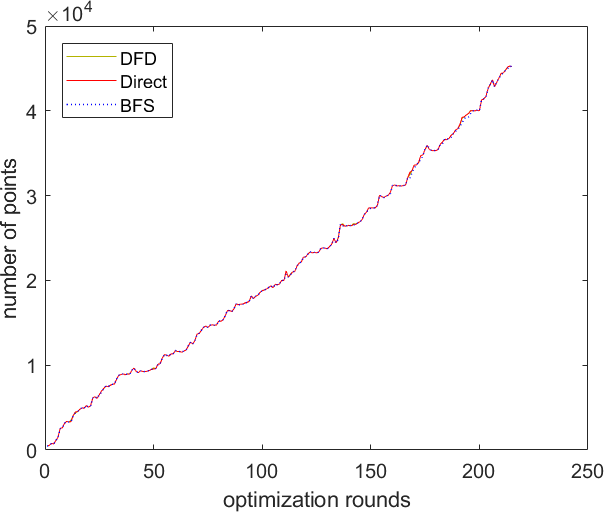}
		}\quad
		\subfigure[Points to be updated for each round of  optimization in the office scene]{
			\includegraphics[width=3.9cm,height=3cm]{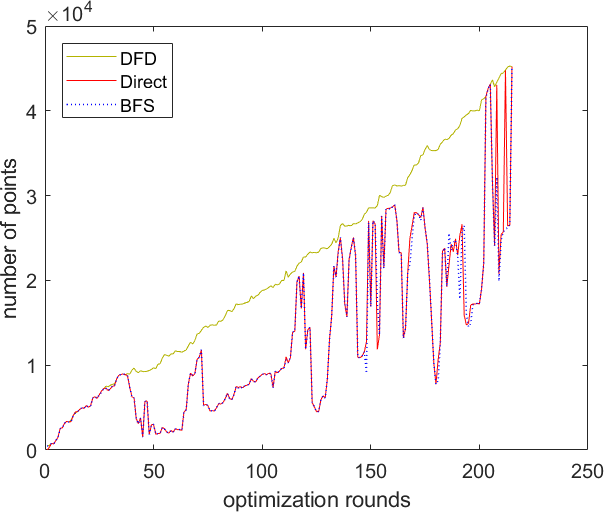}
		}\quad
		\subfigure[Time spent on each round of optimization in the office scene]{
			\includegraphics[width=3.9cm,height=3cm]{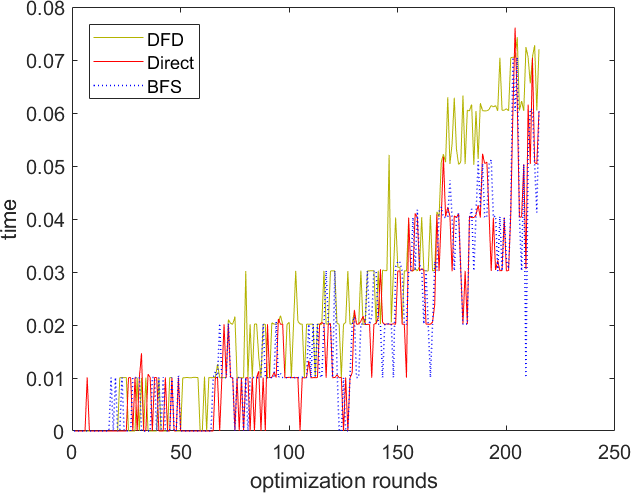}
		} \\
		\subfigure[Deutsches Museum dataset]{
			\includegraphics[width=3.9cm,height=3cm]{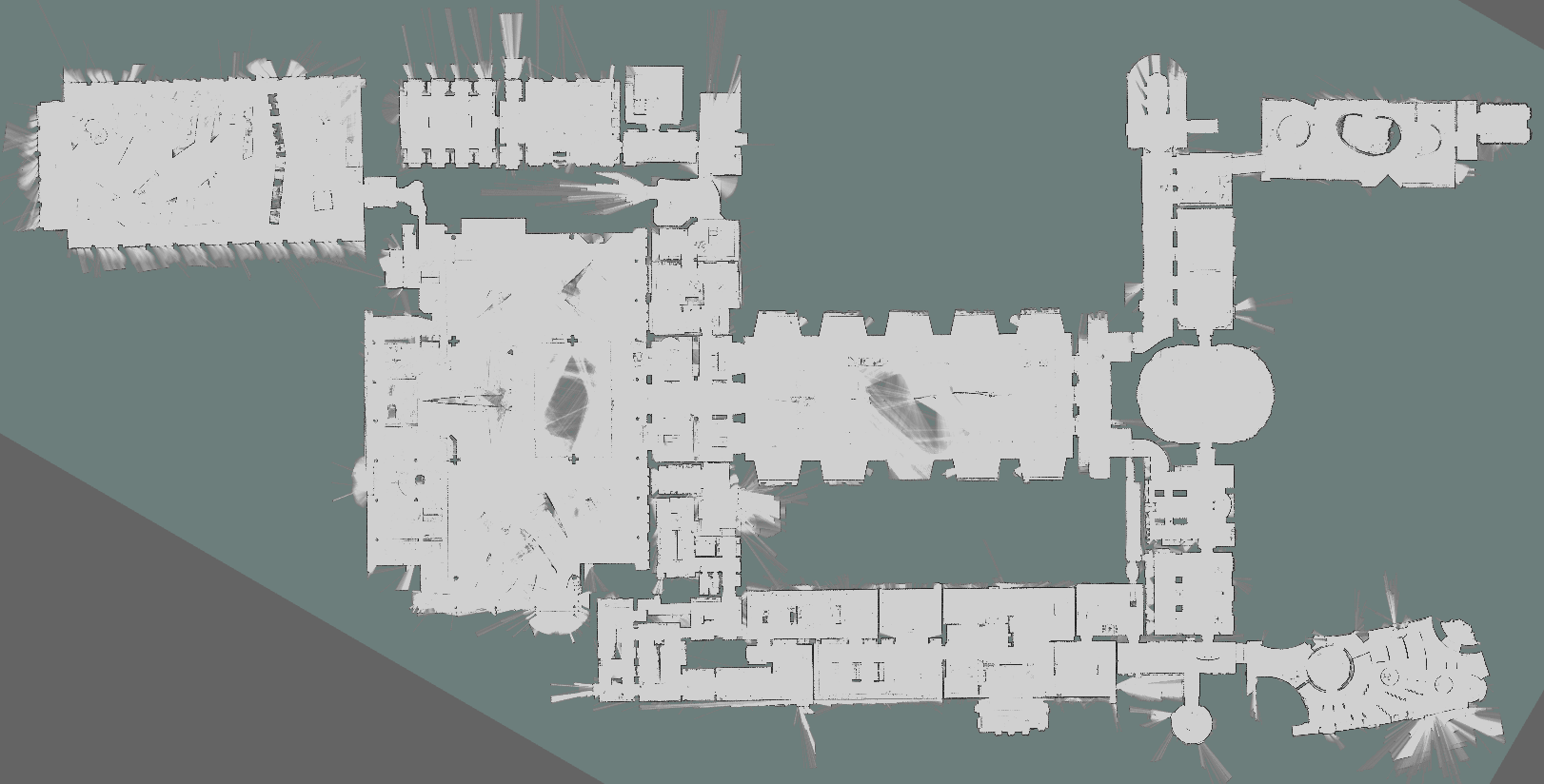}
		}\quad
		\subfigure[Total frontier points of all submaps in the museum scene]{
			\includegraphics[width=3.9cm,height=3cm]{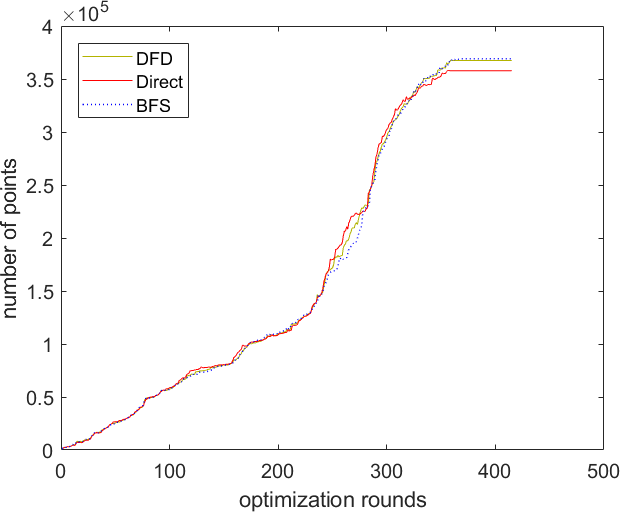}
		}\quad
		\subfigure[Points to be updated for each round of optimization in the museum scene]{
			\includegraphics[width=3.9cm,height=3cm]{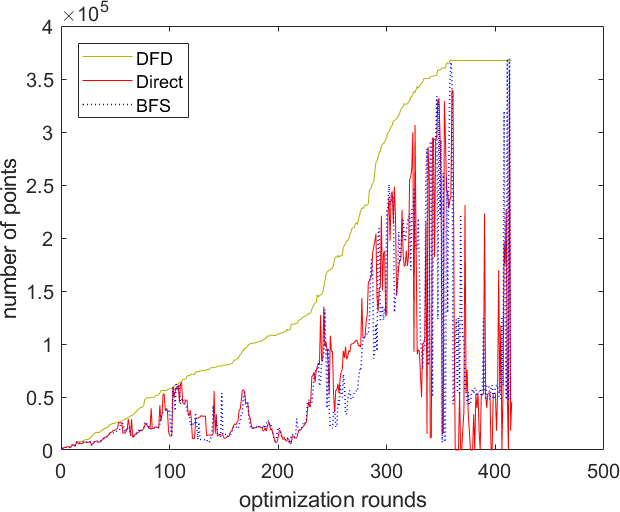}
		}\quad
		\subfigure[Time spent on each round of optimization in the museum scene]{
			\includegraphics[width=3.9cm,height=3cm]{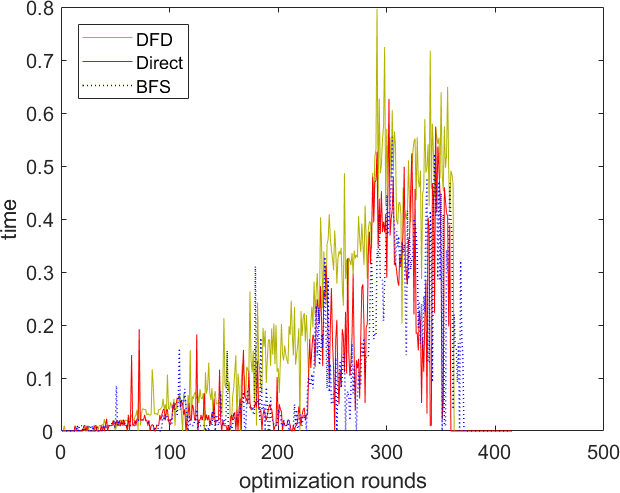}
		}
		\caption{Performance comparison of the three methods in the office and museum scenes.}
		\label{performance}
	\end{figure*}

	We record all data required for robot active exploration in our office scene as a rosbag. 
	Then we test the DFD, the BFS and the Direct methods using our rosbag and the Deutsches Museum dataset, as shown in Fig.\ref{performance}(a)(e). 
	The DFD program code we use is open sourced on GitHub by its author and we use the same parameters in all three methods.
	First, we compare the performance of the three methods in frontier detection. 
	Fig.\ref{performance}(b)(f) show the numbers of all local frontier points detected by the three methods, respectively. It can be seen that there is a slight difference between them. 
	Although the input data is exactly the same, due to the randomness of the algorithms, e.g., the Levenberg–Marquardt algorithm, the submaps created by Cartographer cannot guarantee complete consistency. But the difference between them is small enough. 
	It does not affect the validity of our experiments. 
	
\begin{figure*}
		\subfigure[Frontier accuracy in the office scene]{
			\includegraphics[width=5.5cm,height=3.8cm]{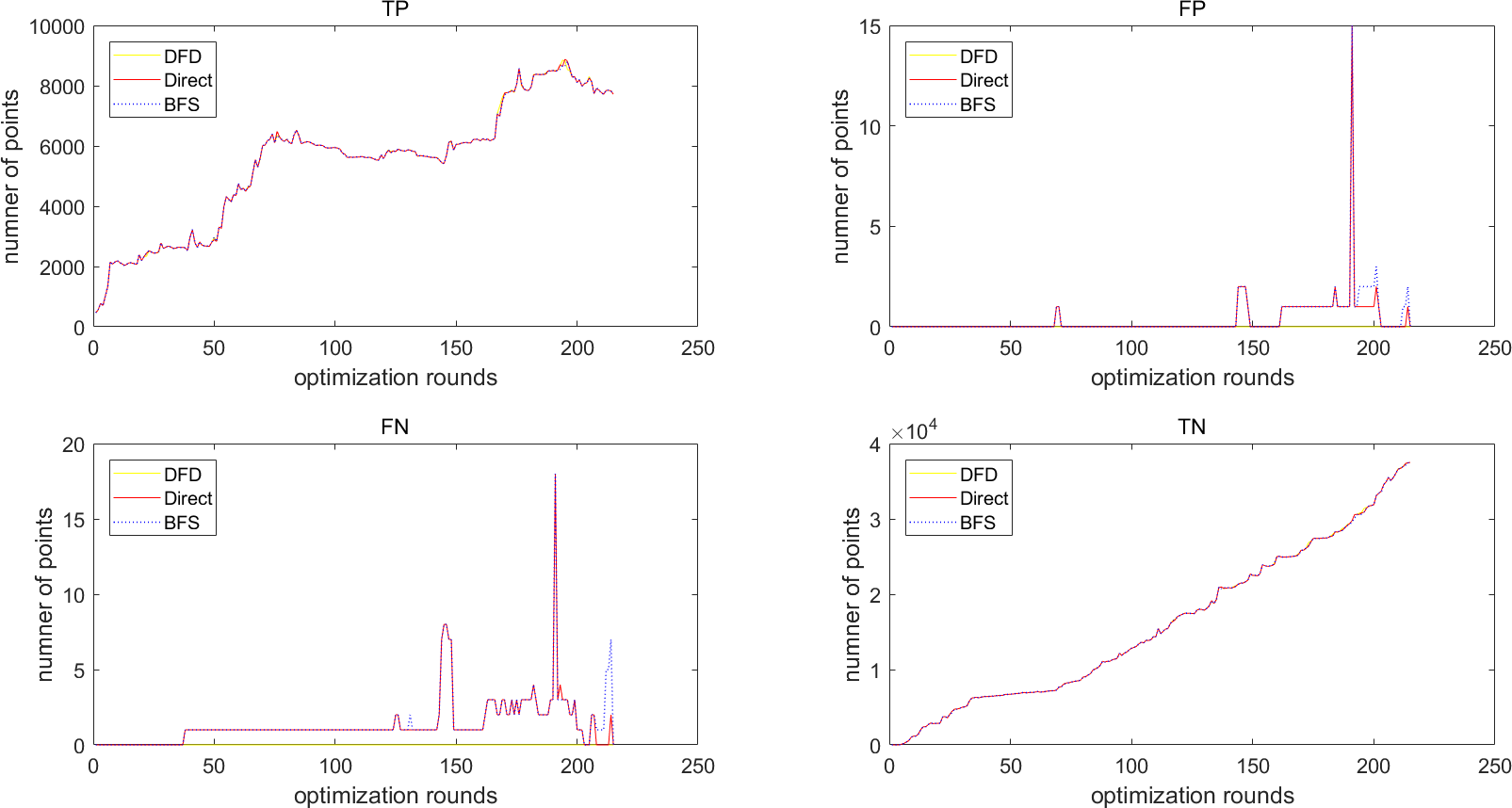}
		}\quad
		\subfigure[Points to be updated with different threshold in the office scene]{
			\includegraphics[width=5.5cm,height=3.8cm]{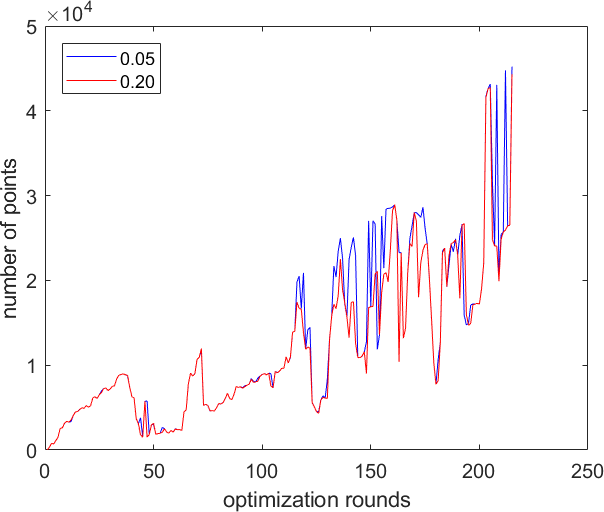}
		}\quad
		\subfigure[Accuracy of the Direct method with different threshold in the office scene]{
			\includegraphics[width=5.5cm,height=3.8cm]{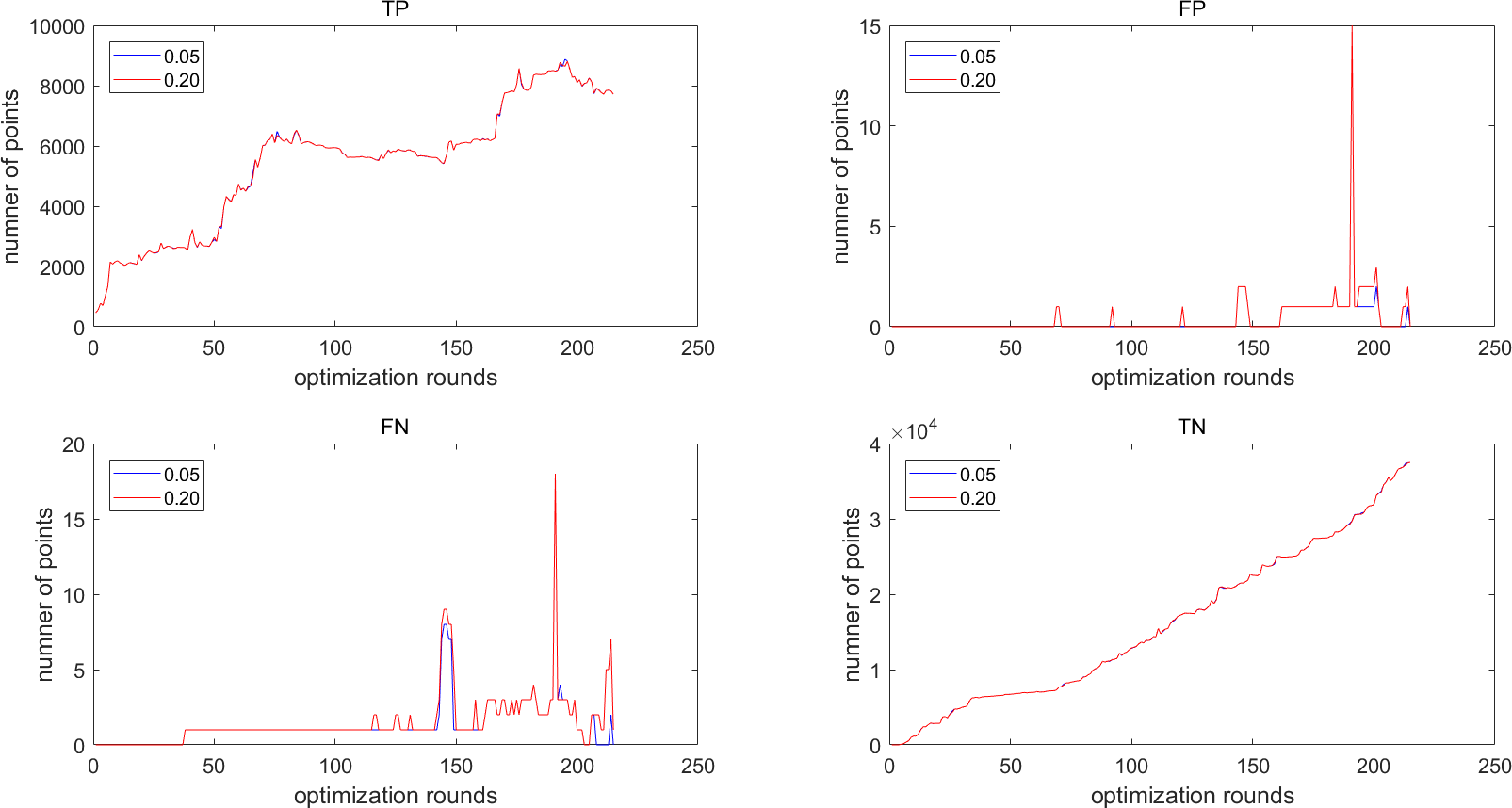}
		} \\
		\subfigure[Frontier accuracy in the museum scene]{
			\includegraphics[width=5.5cm,height=3.8cm]{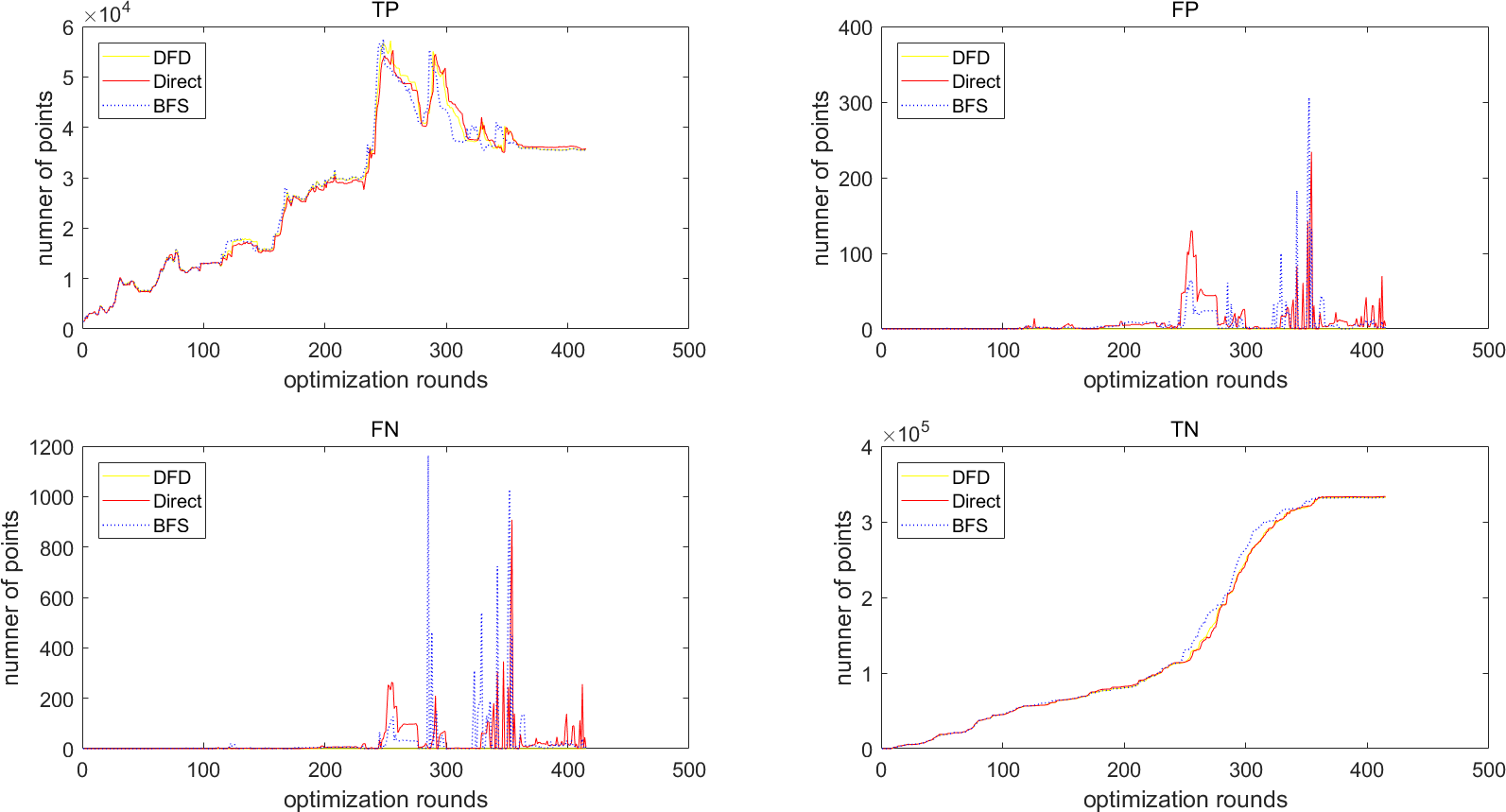}
		} \quad
		\subfigure[Points to be updated with different threshold in the museum scene]{
			\includegraphics[width=5.5cm,height=3.8cm]{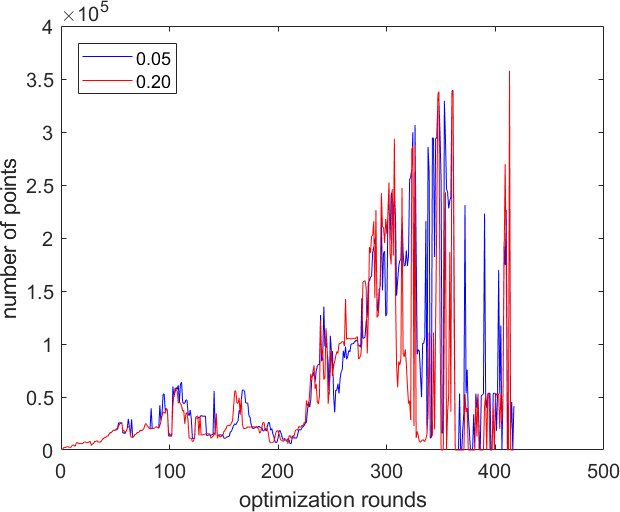}
		}\quad
		\subfigure[Accuracy of the Direct method with different threshold in the museum scene]{
			\includegraphics[width=5.5cm,height=3.8cm]{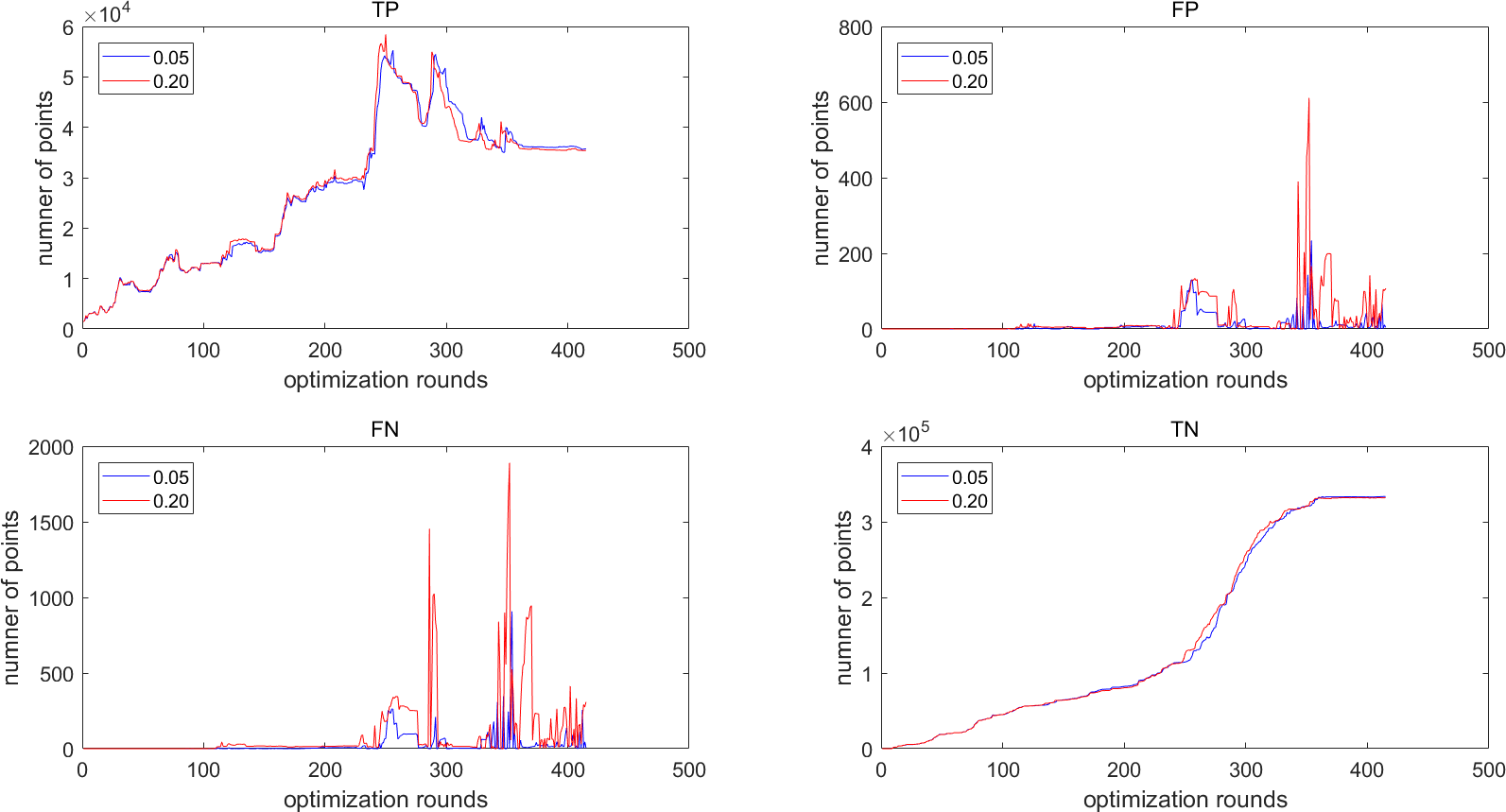}
		}
		\caption{(a)(d) show the accuracy comparison of the three methods in the office and museum scenes. 
			(b)(e) show the performance comparison of different threshold of the DFD method in the two scenes. 
			(c)(f) show the accuracy comparison of different threshold of the DFD method in the two scenes. }
		\label{threshold}
	\end{figure*}

	Fig.\ref{performance}(c)(g) show the number of points that need stabbing query and (d)(h) show the time required for each optimization of the three methods. 
	It can be seen that the calculation and the time required for the BFS and the Direct methods are significantly smaller than those of the DFD method. 
	The Direct method is slightly slower than the BFS method. 
	This is because it records the accumulative shifts, which may query more local frontier points than the BFS method. 
	We reviewed all local frontier points to verify the accuracy of our method, as shown in Fig.\ref{threshold}(a)(d). 
	
	We quantitatively compare the accuracy and performance of the three methods for the two scenes. 
	In the office scene, the accuracy of the Direct and the BFS methods are higher than $99.9\%$. The precision is more than 99.83$\%$ and the recall is more than $99.8\%$. 
	The performance $P$ is improved by $39.3\%$. 
	\begin{equation}
	P = 1-\frac{\sum_{i=0}^{N} Direct_{i}}{\sum_{j=0}^{N} DFD_{j}}
	\end{equation}
	where $N$ represents the number of optimization rounds. The $Direct$ and the $DFD$ represent the number of points that should be updated during each optimization by the two methods, respectively. 
	Thus, we calculated the ratio of the area between the DFD and the Direct curves to the area between the DFD and the $X$ coordinate axis in Fig.\ref{performance}(c). 
	Compared with the Direct method, the performance of the BFS method is $98.79\%$ of the Direct method. 
	However, in some iterations, the BFS method generates a small amount of additional errors. We consider the performance of these two methods to be similar in small scenes. 
	
	In the museum scene, the accuracy of the Direct method is more than $99.72\%$, and the precision being more than $99.4\%$ and the recall being more than $97.75\%$. 
	The performance $P$ of the Direct method is improved by $59.97\%$. 
	As the scene grows, the performance of the Direct method can be significantly improved, but introducing of a very small amount of errors. 
	The performance of the BFS method is $97.4\%$ of the Direct method. 
	Compared with the Direct method, the BFS method has only a small improvement in performance, but it brings a lot of error detection. 
	Therefore, we do not recommend using the BFS method in large scenarios.
	I
	
	We also discuss the effects of different threshold values, as shown in Fig.\ref{threshold}(b,c,e,f). 
	We set the thresholds to $5$cm and $20$cm, respectively. 
	It can be seen that, with the increase of the threshold value, there is only a small improvement in performance, but a large error is introduced accordingly. 
	This is especially noticeable in the large museum scene. 
	
	\section{conclusion}
	
	In this paper, we present an integrated active exploration method for 2D graph SLAM based on efficient frontier detection and robust reachability analysis. 
	Compared with the state-of-the-art frontier detection method in the graph-SLAM, our method uses the feedback information of the graph optimization process to improve the performance of frontier detection.  
	Through reachability analysis of the frontiers and their clusters, we assign a more robust explorable target to the robot, so that active exploration can be applied to real complex scenes with various obstacles. 
	\\
	

	
	
	

	

	\bibliographystyle{IEEEtran}
	\bibliography{bibfile}

\end{document}